\documentclass{article}

\usepackage{microtype}
\usepackage{graphicx}
\usepackage{booktabs} % for professional tables

\usepackage{hyperref}

\usepackage[accepted]{icml2024_tf2m}

\usepackage{amsmath}
\usepackage{amssymb}
\usepackage{mathtools}
\usepackage{amsthm}

\usepackage[capitalize,noabbrev]{cleveref}

\usepackage{bbm}
\usepackage{adjustbox}
\def\1{\mathbbm{1}}

\setcitestyle{numbers,square,sorted}

\usepackage{wrapfig}
\usepackage{subfig}

\newcounter{mycounter} % create a new counter, called 'mycounter'

\usepackage{tcolorbox}

\newcommand{\findingbox}[1]{
\vspace{-4pt}
    \stepcounter{mycounter} % Increment counter
    \begin{tcolorbox}[colframe=black,
                      arc=1pt,
                      boxsep=-2pt,
                      ]
        \noindent{\textbf{\textit{Finding \themycounter.}}} #1 
    \end{tcolorbox}
\vspace{-4pt}
}

\theoremstyle{plain}

\theoremstyle{definition}

\theoremstyle{remark}

\usepackage[textsize=tiny]{todonotes}

\icmltitlerunning{A deeper look at depth pruning of LLMs}

\begin{document}

\twocolumn[
\icmltitle{A deeper look at depth pruning of LLMs}



\icmlsetsymbol{equal}{*}

\begin{icmlauthorlist}
\icmlauthor{Shoaib Ahmed Siddiqui}{cam}
\icmlauthor{Xin Dong}{nvidia}
\icmlauthor{Greg Heinrich}{nvidia}
\icmlauthor{Thomas Breuel}{nvidia}
\icmlauthor{Jan Kautz}{nvidia}
\icmlauthor{David Krueger}{cam}
\icmlauthor{Pavlo Molchanov}{nvidia}
\end{icmlauthorlist}

\icmlaffiliation{cam}{University of Cambridge, UK}
\icmlaffiliation{nvidia}{NVIDIA Research, USA}

\icmlcorrespondingauthor{Shoaib Ahmed Siddiqui}{msas3@cam.ac.uk}
% \icmlcorrespondingauthor{Pavlo Molchanov}{pmolchanov@nvidia.com}

\icmlkeywords{Depth pruning, LLMs, efficiency.}

\vskip 0.3in
]



\printAffiliationsAndNotice{}  % leave blank if no need to mention equal contribution

\begin{abstract}
Large Language Models (LLMs) are not only resource-intensive to train but even more costly to deploy in production.
Therefore, recent work has attempted to prune blocks of LLMs based on cheap proxies for estimating block importance, effectively removing 10\% of blocks in well-trained LLaMa-2 and Mistral 7b models without any significant degradation in downstream metrics.
This work explores different block importance metrics by considering adaptive metrics such as Shapley value in addition to static ones explored in prior work.
We show that \textit{adaptive metrics exhibit a trade-off in performance between tasks i.e., improvement on one task may degrade performance on the other due to differences in the computed block influences}.
Furthermore, we extend this analysis from a complete block to individual self-attention and feed-forward layers, highlighting the propensity of the self-attention layers to be more amenable to pruning, even allowing \textit{\textbf{removal of up to 33\% of the self-attention layers without incurring any performance degradation on MMLU for Mistral 7b}} (significant reduction in costly maintenance of KV-cache).
Finally, we look at simple performance recovery techniques to emulate the pruned layers by training lightweight additive bias or low-rank linear adapters.
% that can be trained efficiently even on a single GPU.
\textit{Performance recovery using emulated updates avoids performance degradation for the initial blocks (up to 5\% absolute improvement on MMLU)}, which is either competitive or superior to the learning-based technique\footnote{Code to reproduce experiments is publicly available: \url{https://github.com/shoaibahmed/llm_depth_pruning}.}.
% The code to reproduce our experiments is publicly available: \url{https://github.com/shoaibahmed/llm_depth_pruning}.
% Code: \url{https://github.com/shoaibahmed/llm_depth_pruning}.
\end{abstract}
\vspace{-8mm}

% \footnote{Code: \url{https://github.com/shoaibahmed/llm_depth_pruning}.}

% \input{sections/abstract}
% \input{sections/intro}

\section{Introduction}

\begin{figure}[t]
    \centering
    \includegraphics[width=0.48\textwidth]{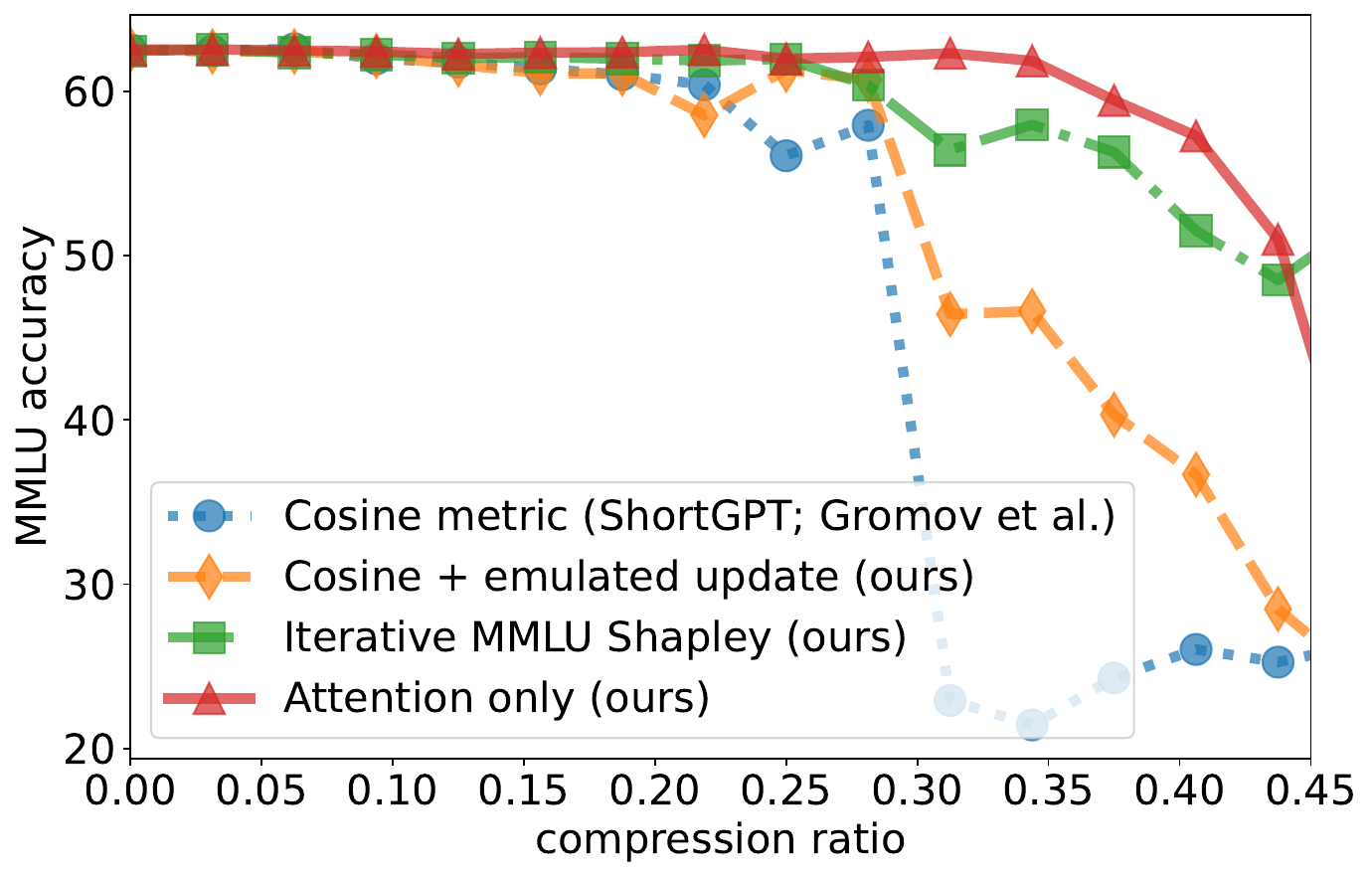}
    \vspace{-7mm}
    \caption{\textbf{Our self-attention pruning, adaptive metrics, as well as emulated updates in comparison to cosine block pruning} visualized w.r.t. compression ratios on Mistral 7b (see Fig.~\ref{fig:main_llama} for results on LLaMa-2 7b).}
    \label{fig:main_mistral}
    \vspace{-5mm}
\end{figure}

The utility of training language models with an increasing number of parameters is currently undisputed~\citep{kaplan2020scalinglaws,brown2020gpt3,touvron2023llama2,jiang2023mistral}.
This has led to tremendous gains in performance, including the emergence of in-context learning~\citep{brown2020gpt3}.
However, aside from training, which is a one-time cost, deploying these models in production presents unique challenges, particularly high inference costs.
Therefore, efficiency research for language models has gained significant popularity in the recent past~\citep{frantar2022gptq,ashkboos2024slicegpt,van2023llmsurgeon,men2024shortgpt}.

An extreme form of (structured) pruning is full-block pruning from a pretrained model. \citet{men2024shortgpt,gromov2024unreasonable} recently showed that it is possible to completely drop blocks from a range of pretrained language models based on cheap proxies for block importance/influence such as computing the cosine distance between the input and the output representations for each block in a transformer. 

This paper explores this direction further by analyzing the impact of different metrics on block identification, specifically focusing on adaptive metrics such as the one based on Shapley-value~\citep{shapley1953value} in addition to static ones such as cosine block influence mainly used in prior work~\citep{men2024shortgpt,gromov2024unreasonable}.
We further look at the impact of individual self-attention and the feed-forward layers, which together form a block as analyzed by prior work.
Finally, we evaluate the effectiveness of simple strategies for performance recovery, including a simple baseline of an additive bias (called `emulated update`) which is based on the empirical mean of the update applied by the block, as well as learning-based techniques such as the training of a low-rank adapter, following prior literature on layer-stitching~\citep{bansal2021layerstitching}.

We visualize our main findings in Fig.~\ref{fig:main_mistral} where `block cosine` represents the method employed in prior work~\citep{men2024shortgpt,gromov2024unreasonable}.
Note that the x-axis represents the compression ratio computed based on the total number of layers. Therefore, it does not represent an equal number of parameters being pruned as a block comprises both a self-attention and a feed-forward layer.
The figure highlights that the model can tolerate a higher fraction of pruning for self-attention layers in comparison to feed-forward layers or complete model blocks (computed using loss-shapley block influence -- see Fig.~\ref{fig:attn_vs_mlp} for a direct comparison).
Pruning self-attention layers can provide a significant boost in model efficiency due to the costly maintenance of KV-cache at inference time.
We further include an upper bound on MMLU performance by iteratively computing Shapley-value-based block influence directly on the MMLU test set to understand the possible improvements achievable when leveraging information about the task.
% Due to this inference cost, KV-cache reduction has been a significant topic of attention in the recent past~\citep{xiao2023streaminglm,nawrot2024dynamic}.
Our main findings can be summarized as:

\findingbox{
We highlight the impact of different block influence metrics on downstream model performance (Fig.~\ref{fig:metrics_comparison_simple}).
Our analysis reveals a trade-off in performance between tasks when evaluating adaptive metrics e.g., Shapley-value~\citep{shapley1953value} (Fig.~\ref{fig:metrics_comparison}). Furthermore, we find that performance on some tasks degrades significantly, such as GSM-8k~\citep{cobbe2021gsm8k} even after pruning a single block from the model (Fig.~\ref{fig:llama2_harness} and Fig.~\ref{fig:mistral_harness}).}
\vspace{1mm}
\findingbox{We further dissect a block into attention and feed-forward layers to separately evaluate their influence, and highlight a higher propensity to prune attention layers in contrast to feed-forward layers while preserving MMLU accuracy (Fig.~\ref{fig:attn_vs_mlp}).}
\vspace{1mm}
\findingbox{We evaluate two simple performance recovery techniques: (i) simple empirical mean of the block update, and (ii) training of a low-rank linear adapter in place of the missing block (Fig.~\ref{fig:performance_recovery_simple}). Surprisingly, our results show that applying a simple average block update is either competitive or superior in performance as compared to the learning-based approach potentially due to overfitting and catastrophic forgetting.
% In particular, we observe performance improvements by attaching a low-rank linear adapter for LLaMa-2 (Fig.~\ref{fig:linear_adapters_rank_8_llama2_relative}), but performance instead degrades for Mistral 7b (Fig.~\ref{fig:linear_adapters_rank_8_mistral_relative}) which is aligned with the results on model healing from~\citep{gromov2024unreasonable}.
}

\section{Related work}
\label{sec:related_work}

\citet{he2016resnets} popularized residual layers for training deep models. \citet{veit2016resnetsasensembles} hypothesized residual networks to be ensembles of smaller sub-networks, exhibiting layer dropping at inference time with minimal performance degradation. \citet{veit2018convaig} extended this to dynamic input-conditioned layer-skipping.
\citet{vaswani2017transformer} introduced the famous transformer architecture by combining ideas of attention and residual networks to develop highly performant architectures.
While there is a large body of work on pruning for efficient inference, we focus on depth pruning and refer the readers to \citet{wan2023efficientllms} for a more comprehensive treatment of the literature.

\citet{samragh2023weightcloning} initialized a subnetwork using blocks from a pretrained GPT-2~\citep{radford2019gpt2}.
Sheared LLaMa~\citep{xia2023shearedllama} proposed an optimization-based view for simultaneous depth and width pruning of LLaMa-2 7b model~\citep{touvron2023llama2}.
% They also utilized extensive fine-tuning of a low-rank adapter as a performance recovery technique.
Shortened LLaMa~\citep{kim2024shortenedllama} showed that depth pruning is competitive against width-only pruning, or a sophisticated combination of both, while exploring several different metrics for estimating block influence.
% They constructed proxy metrics for block influence estimation based on the weights of the network, as well as the change in the downstream metric.
% One of the considered metrics was based on the drop in perplexity, which resembles our loss-shapley on the language modeling task.
\citet{men2024shortgpt} focused on LLaMa-2~\citep{touvron2023llama2} by using cosine distance (between activations before and after a block) as a proxy of block importance.
\citet{gromov2024unreasonable} also showed similar results on both LLaMa-2~\citep{touvron2023llama2} and Mistral~\citep{jiang2023mistral}, while also proposing a healing method by optimizing low-rank adapters~\citep{hu2021lora} for the remaining blocks. This healing process is aimed at minimizing performance degradation with block pruning.

% FFN-skip ref: https://arxiv.org/pdf/2404.03865.pdf
\citet{jaiswal2024ffnskipllm} extended this by computing the cosine similarity of the representations at inference time and skipping only the feed-forward layers in the less important regions of the network (particularly in the middle of the network).
In a similar spirit, \citet{raposo2024mixtureofdepths} trained a router to decide which blocks to skip i.e., reduce network depth (similar to the expert router in MoE~\citep{shazeer2017mixtureofexperts}).

This paper attempts to take a deeper look at depth pruning of LLMs by looking at multiple metrics, datasets, block granularities (going to individual feed-forward and self-attention layers), and recovery techniques to establish the utility of each of these decisions on the resulting model.

\section{Methods}

% We consider a regular transformer architecture with self-attention, feed-forward layers, and input layer-norms~\citep{vaswani2017transformer}.

\subsection{Block influence metrics}
\label{subsec:block_inf_metrics}

We consider different block influence metrics including cosine that computes the angular distance between the input and output representations to a block~\citep{men2024shortgpt,gromov2024unreasonable}, relative $L_1/L_2$ that computes the norm of the update with respect to the norm of the input representation~\citep{samragh2023weightcloning,men2024shortgpt}, and Shapley-value-based~\citep{shapley1953value} estimate which computes the marginal contribution of a block by computing the difference in performance between all subsets where a particular block is present and the block is absent.
We focus on computing Shapley-value for the distance to the full model logits or the language modeling loss, except Fig.~\ref{fig:main_mistral} where we compute it directly for the 0-1 loss on the MMLU test set. In this case, the Shapley computation is:
\begin{equation*}
    \textsc{SHAP}^{l} = \mathbb{E}_{\mathbf{x}, s \subseteq S \setminus \{l\}} \left[\mathcal{L}(O^{s \cup \{l\}}(\mathbf{x})) - \mathcal{L}(O^{s}(\mathbf{x}))\right]
\end{equation*}
\noindent where $S$ represents the full set of blocks, $s$ represents a subset of the blocks, and $O^{s}(\mathbf{x})$ represents the output of the model on input $\mathbf{x}$ when using a subset of the blocks $s$.

\subsection{Performance recovery techniques}
\label{subsec:performance_recovery_techniques}

As dropping blocks can induce a distribution shift in the representation, we consider two simple performance recovery techniques.

\subsubsection{Emulated update}

Emulated update is a particularly simple strategy that computes the average update applied by each block on a small calibration set, and applies that average additive update at inference time when a particular block is dropped.
This can be viewed as having a `bias` term instead of the full block.

\subsubsection{Low-rank linear adapters}
\label{subsec:linear_adapters}

Following ideas from layer-stitching literature~\citep{bansal2021layerstitching}, we evaluate the efficacy of training low-rank linear adapters as a performance recovery measure. More precisely, we introduce two low-rank matrices per block to be learned (aside from the per-channel weightings learned as part of the normalization layer):
This is similar in spirit to the healing process based on LoRA adapters by~\citet{gromov2024unreasonable}, but we apply it in place of the missing blocks instead of applying it on the remaining blocks.

We consider three different ways of training these linear adapters: (i) minimize MSE between the output of the block and the low-rank adapter, (ii) use supervised fine-tuning (SFT) on the final output, and (iii) use logit-distillation on the final model output.
We use the calibration set to train this low-rank adapter independently one block at a time.

\section{Experiments}

We use a subset of $150k$ sequences (context window of 2048 tokens) with cross-document concatenation from OpenWebText~\citep{openwebtext} as our full calibration set, and another $15k$ sequences as our validation set.
We report 5-shot MMLU~\citep{hendrycks2020mmlu} accuracy as commonly reported in the literature~\citep{men2024shortgpt,gromov2024unreasonable}.
We use LLaMa-2 7b~\citep{touvron2023llama2} and Mistral 7b~\citep{jiang2023mistral} for our evaluations.
We focus on 7b models primarily due to them being a sweet spot between compute cost and model quality.
Larger models are more amenable to block pruning, as already established in prior work~\citep{men2024shortgpt,gromov2024unreasonable}.
Therefore, we expect our findings to be equally applicable to larger models.

\subsection{Block influence metrics}

\begin{figure}[t]
    \centering
    \includegraphics[width=\linewidth]{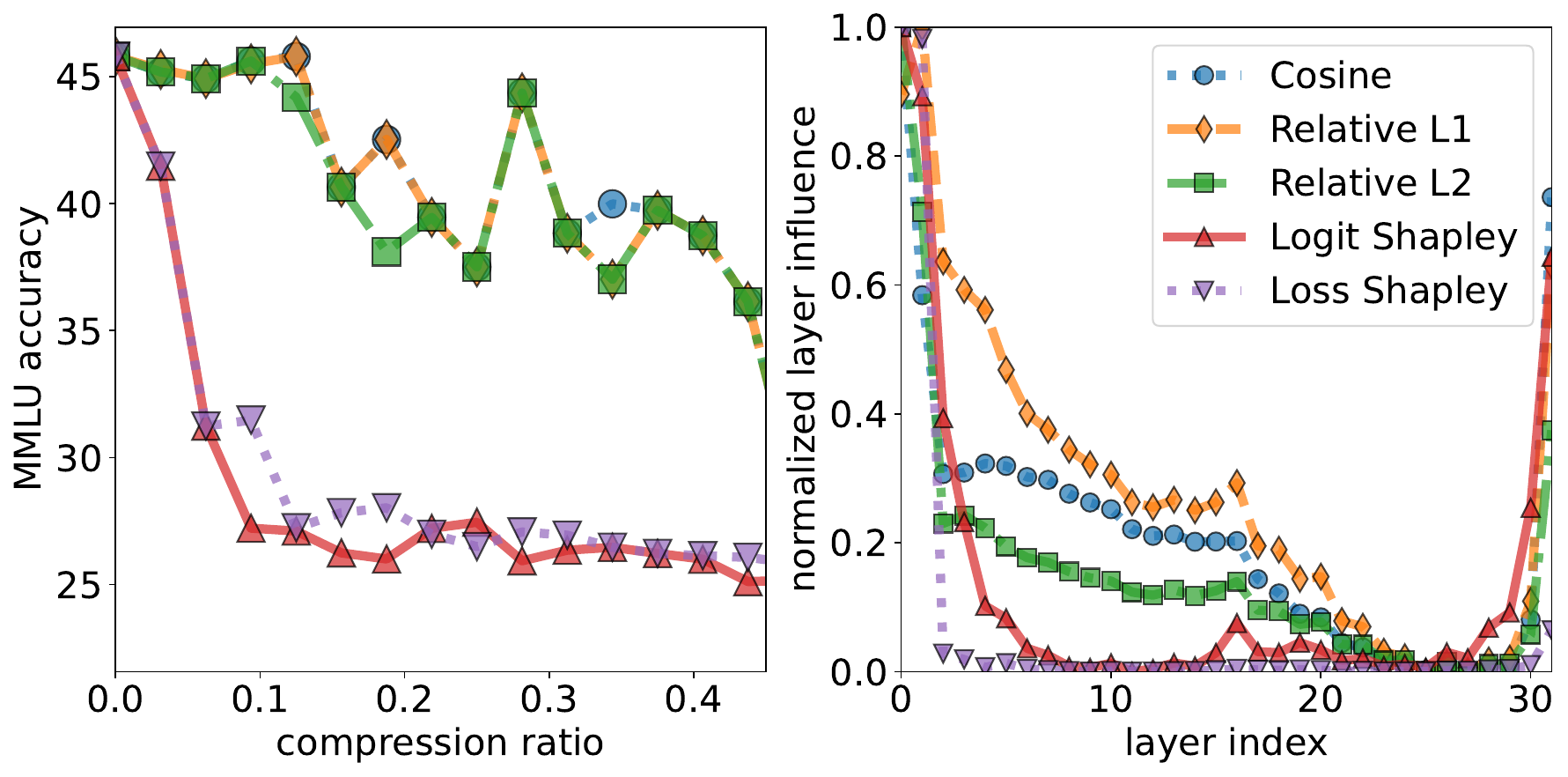}
    \includegraphics[width=\linewidth]{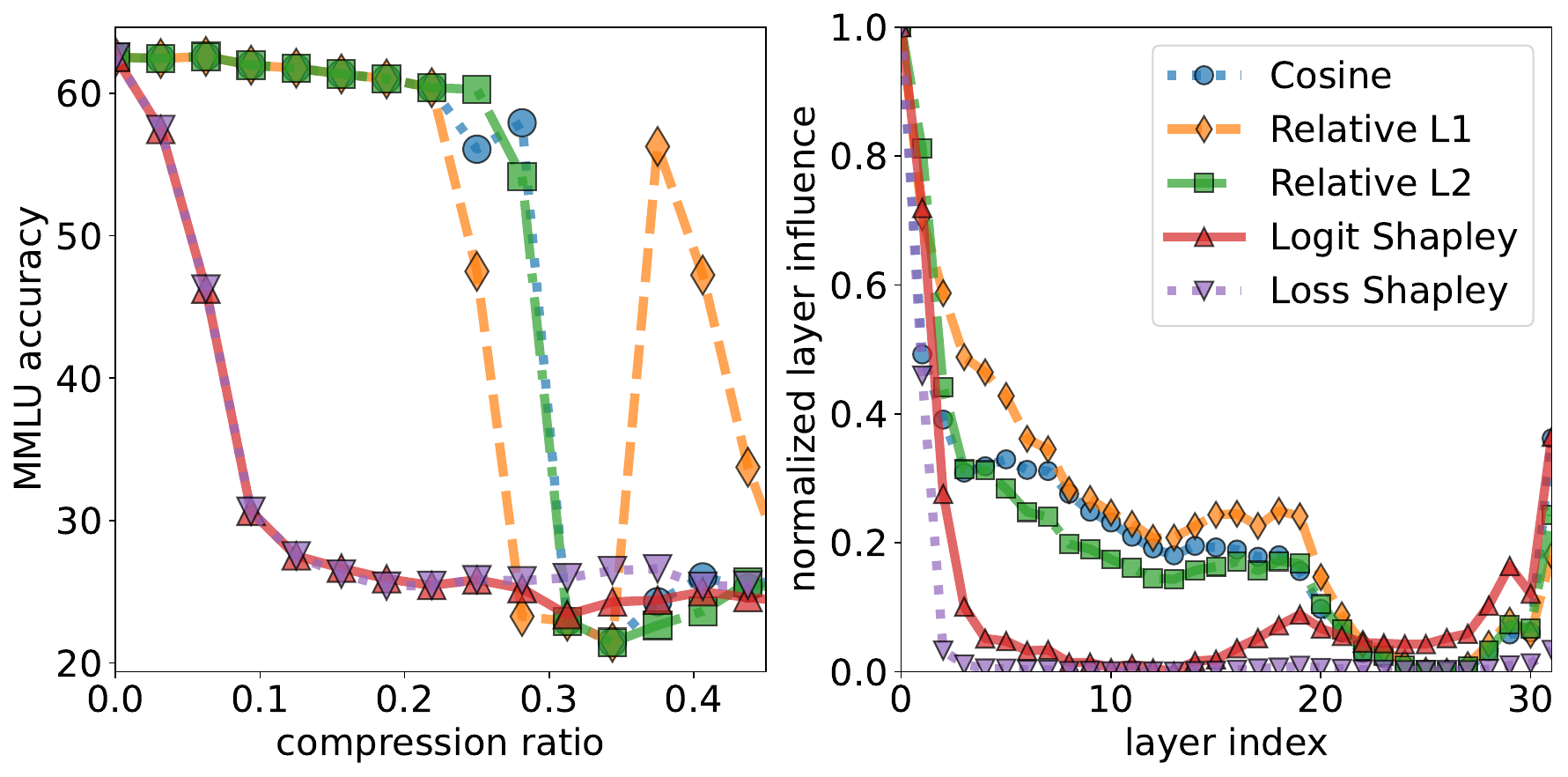}
    \vspace{-6mm}
    \caption{\textbf{Comparison of different block influence metrics} used for block pruning and their impact on downstream performance in terms of MMLU accuracy on LLaMa-2 7b (top) and Mistral 7b (bottom).}
    \label{fig:metrics_comparison_simple}
    \vspace{-5mm}
\end{figure}

We compare different block influence metrics in Fig.~\ref{fig:metrics_comparison_simple} (see Fig.~\ref{fig:metrics_comparison} for more detailed results), where we apply min-max normalization on the block influences for visualization.
As cosine distance-based block influence matches the proposed approach in~\citep{men2024shortgpt,gromov2024unreasonable}, our results also closely match theirs in terms of robustness against block pruning.
Furthermore, the blocks that are assigned the least importance are located towards the end of the network i.e., the second half.
Relative $L_{p}$ norm metrics also closely follow this trend, and hence, achieve robustness comparable to that of cosine distance.
% ~\citep{gromov2024unreasonable}.

A significant deviation however is shown by the Shapley-value-based estimation, which is an adaptive metric, providing significant gains in terms of reduction of the average loss (visualized in Fig.~\ref{fig:metrics_comparison}) as it is specifically computed to reduce the model's language modeling loss.
However, this results in a drastic reduction in accuracy when evaluating MMLU.
Block influence plots show that in contrast to all other metrics that primarily focus on the second half of the network for removal, Shapley instead focuses on removing blocks from the first half of the network.
When computing loss-based Shapley-value directly on the MMLU test set, we see that performance is significantly better than the cosine baseline (see Fig.~\ref{fig:main_mistral}), highlighting that task-specific (Shapley-based) block dropping might retain higher performance at the same number of dropped blocks as compared to task-agnostic techniques but at the expense of sacrificing more performance on tasks not directly considered.

We also visualize the results on other tasks from OpenLLM leaderboard~\citep{open-llm-leaderboard} using lm-eval-harness~\citep{eval-harness} in Fig.~\ref{fig:llama2_harness} and Fig.~\ref{fig:mistral_harness}.
These results highlight that despite just a relatively minor impact on performance for MMLU, some metrics are significantly more impeded (such as GSM-8k).
The proxy used in this case i.e., `cosine`, matches closely the right influence values to optimize MMLU, which is not aligned with the importance ideal for other datasets.
On the other hand, adaptive metrics can be optimized individually for each task at hand.

\subsection{Disentangling the impact of individual layers}

\begin{figure}[t]
    \includegraphics[width=\linewidth]{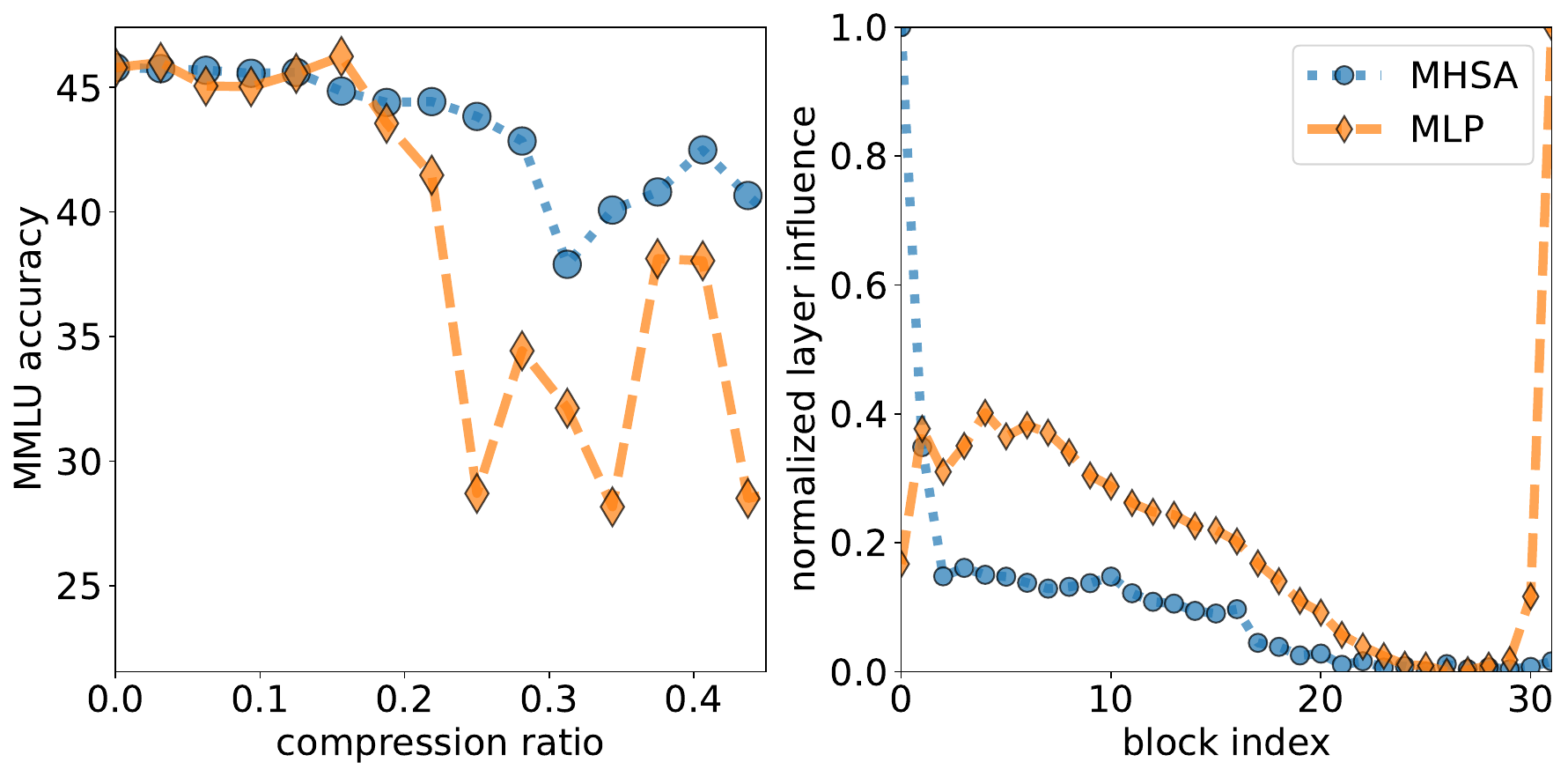}
    \includegraphics[width=\linewidth]{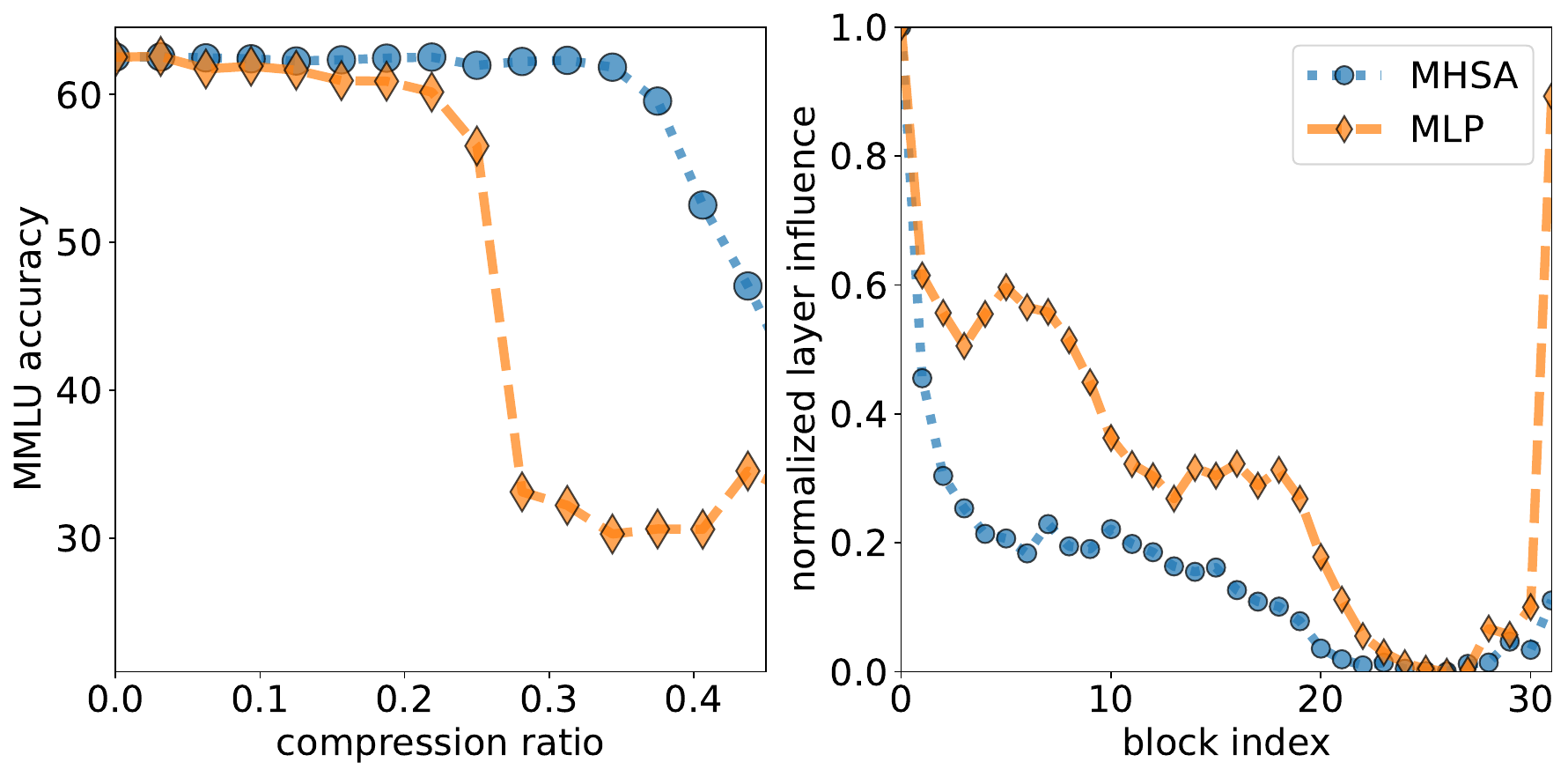}
    \vspace{-6mm}
    \caption{\textbf{Comparison of pruning the self-attention and feed-forward layers of a network using cosine metric} on LLaMa-2 7b (top) and Mistral 7b (bottom), highlighting a higher propensity for the self-attention layers to be pruned.}
    \vspace{-6mm}
    \label{fig:attn_vs_mlp}
\end{figure}

As each block in a transformer is composed of self-attention as well as a feed-forward network, one can apply the same influence techniques to understand the impact of performance when pruning individual layers rather than complete blocks.
The results are highlighted in Fig.~\ref{fig:attn_vs_mlp} (see Fig.~\ref{fig:layer_influence_llama2} and Fig.~\ref{fig:layer_influence_mistral} for complete results where we also visualize the results with joint layer ranking).

We see that models exhibit higher resilience against the dropping of self-attention layers in contrast to feed-forward layers.
In contrast to all other experiments, loss shapley in the case of self-attention layers achieves competitive influence as compared to other approaches, highlighting that it might be easier to estimate in contrast to feed-forward layers (Fig.~\ref{fig:layer_influence_llama2} and Fig.~\ref{fig:layer_influence_mistral}).
Furthermore, Mistral 7b exhibits a higher propensity for self-attention layer removal compared to LLaMa-2 7b.

Although most transformer parameters are in the feed-forward layers, pruning the self-attention layer is equally useful due to its quadratic dependency on sequence length and the complex management of key-value caches, which introduces significant latency~\citep{xiao2023streaminglm,nawrot2024dynamic}. Developing techniques that combine block and layer pruning, potentially pruning only a single layer in some blocks, is an exciting direction for future research.

\subsection{Performance recovery techniques}

\begin{figure}[t]
    \centering
    \includegraphics[width=0.49\linewidth]{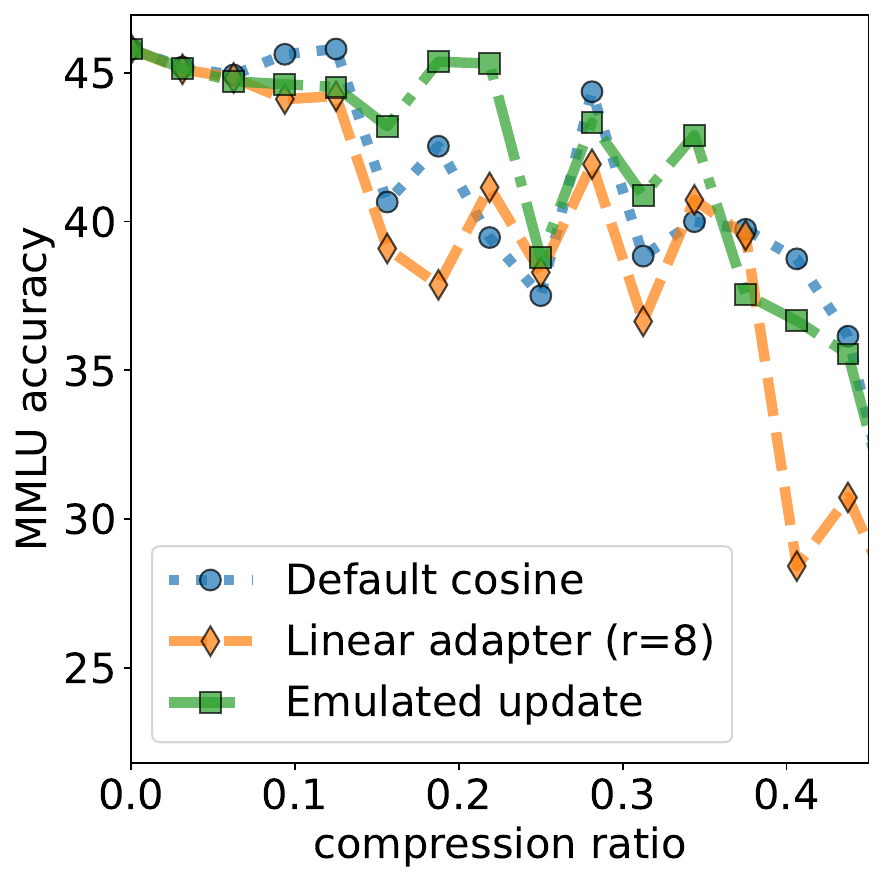}
    \includegraphics[width=0.49\linewidth]{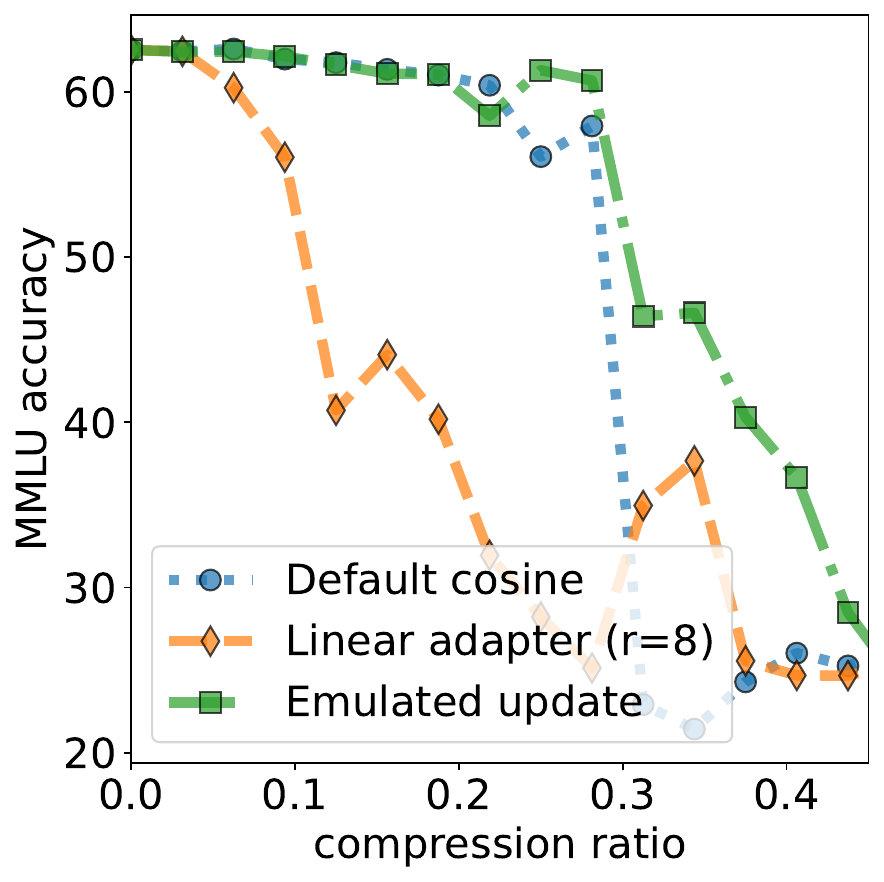}
    \vspace{-3mm}
    \caption{\textbf{Impact on using performance recovery techniques with cosine block influence} on LLaMa-2 7b (left) and Mistral 7b (right). The linear adapter was trained using logit-distillation with a rank of 8.}
    \label{fig:performance_recovery_simple}
    \vspace{-5mm}
\end{figure}

We trained all our adapters for 800 steps with an effective batch size of 8, where the training loss plateaued.
The results for emulated update and linear adapter with a rank of 8 trained using logit-distillation are shown in Fig.~\ref{fig:performance_recovery_simple} (see Appendix~\ref{appendix:low_rank_results} for full results with different ranks and training strategies).
We observe a clear improvement for both LLaMa-2 and Mistral when using emulated update where performance improves significantly at some stages ($\ge 5\%$ on MMLU).
However, performance with a linear adapter is either comparable (LLaMa-2 7b) or worse (Mistral 7b) than the performance improvement observed by the simple emulated block update.
This can be partially attributed to the overfitting of the adapter on the training corpus, which is not reflective of the true pertaining distribution.
\citet{gromov2024unreasonable} observed a similar performance degradation when trying to do parameter-efficient fine-tuning after dropping blocks from Mistral as compared to LLaMa-2.

These results indicate minor differences introduced by later blocks can be recovered using simple techniques. However, this mitigation technique is unable to change the tipping point in performance.

\section{Conclusion}

This paper explores depth pruning of LLMs, highlighting performance differences with various influence techniques, including adaptive ones. Block-sensitivity metrics like Shapley improve perplexity but degrade performance on tasks like MMLU, showing block relevance tension. Our results indicate self-attention layers are more amenable to pruning than feed-forward layers, with performance unaffected by pruning many self-attention layers. We also address misalignment from block pruning using basic performance recovery techniques. Our simplest baseline, using an average update, matches or outperforms low-rank adapters with different training formulations.

\bibliography{main}
\bibliographystyle{icml2024}

%%%%%%%%%%%%%%%%%%%%%%%%%%%%%%%%%%%%%%%%%%%%%%%%%%%%%%%%%%%%%%%%%%%%%%%%%%%%%%%
%%%%%%%%%%%%%%%%%%%%%%%%%%%%%%%%%%%%%%%%%%%%%%%%%%%%%%%%%%%%%%%%%%%%%%%%%%%%%%%
% APPENDIX
%%%%%%%%%%%%%%%%%%%%%%%%%%%%%%%%%%%%%%%%%%%%%%%%%%%%%%%%%%%%%%%%%%%%%%%%%%%%%%%
%%%%%%%%%%%%%%%%%%%%%%%%%%%%%%%%%%%%%%%%%%%%%%%%%%%%%%%%%%%%%%%%%%%%%%%%%%%%%%%
\newpage
\appendix
\onecolumn

\begin{figure}[t]
    \centering
    \includegraphics[width=0.48\textwidth]{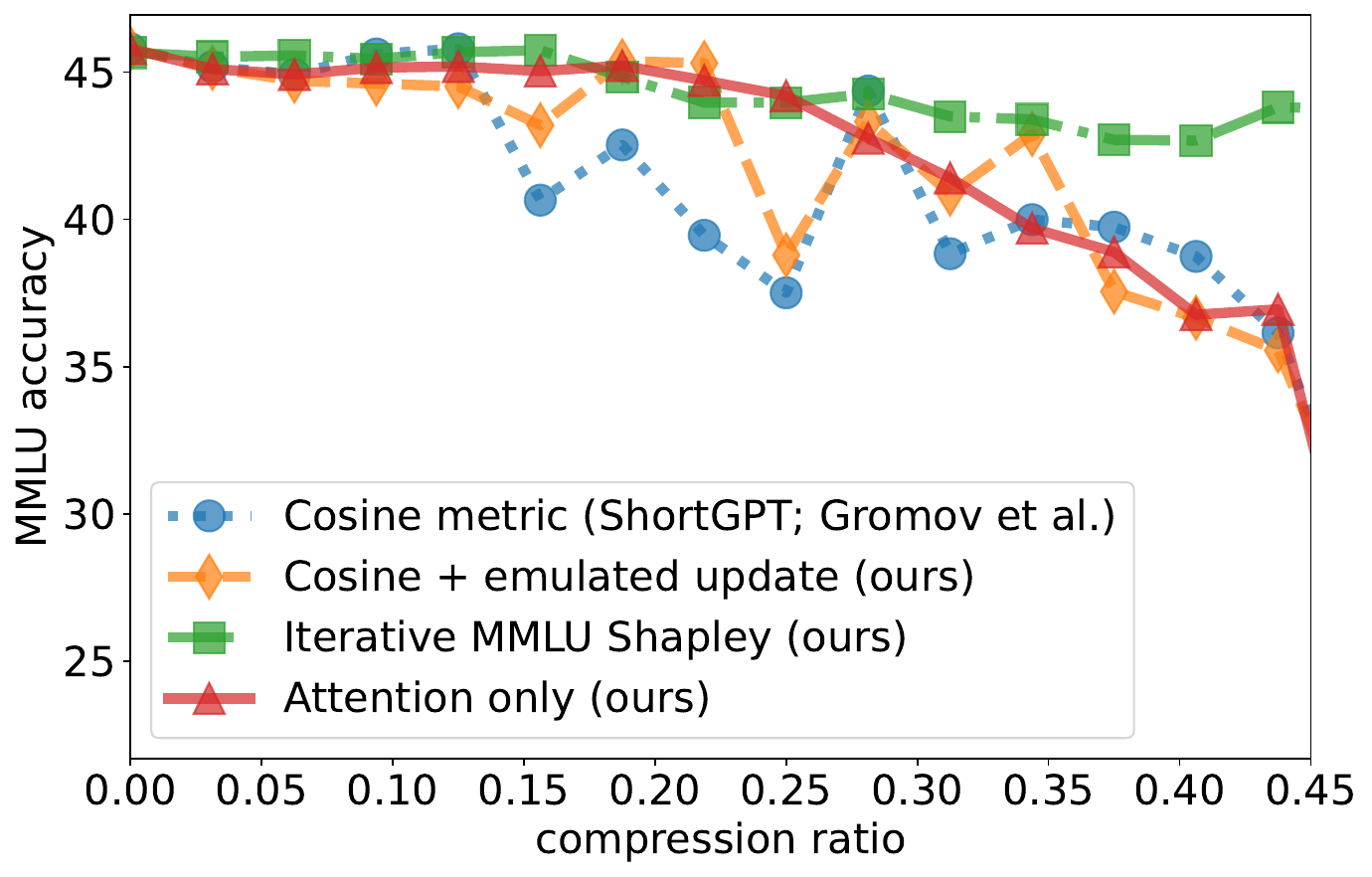}
    \vspace{-3mm}
    \caption{\textbf{Our self-attention pruning, adaptive metrics, as well as emulated updates in comparison to cosine block pruning} visualized w.r.t. compression ratios on LLaMa-2 7b (see Fig.~\ref{fig:main_mistral} for results on Mistral 7b).}
    \label{fig:main_llama}
    \vspace{-5mm}
\end{figure}

\section{Results on other datasets}

\begin{figure}[t]
    \centering
    \includegraphics[width=\textwidth]{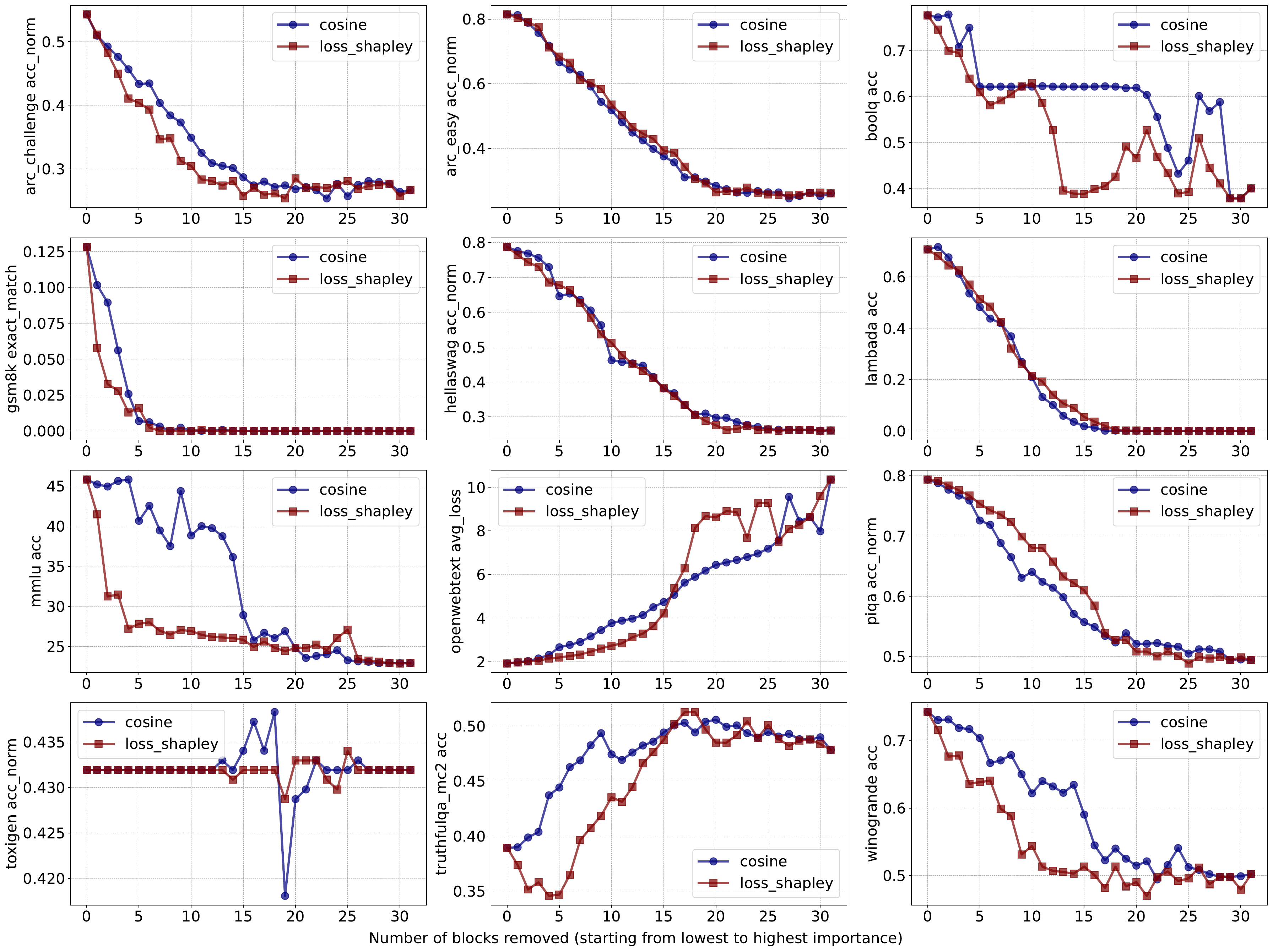}
    \caption{\textbf{LLaMa-2 7b evaluation on multiple datasets from eval-harness~\citep{eval-harness}}, highlighting the impact of pruning on some tasks is more significant (such as GSM-8k~\citep{cobbe2021gsm8k}) as compared to others.}
    \label{fig:llama2_harness}
\end{figure}

\begin{figure}[t]
    \centering
    \includegraphics[width=\textwidth]{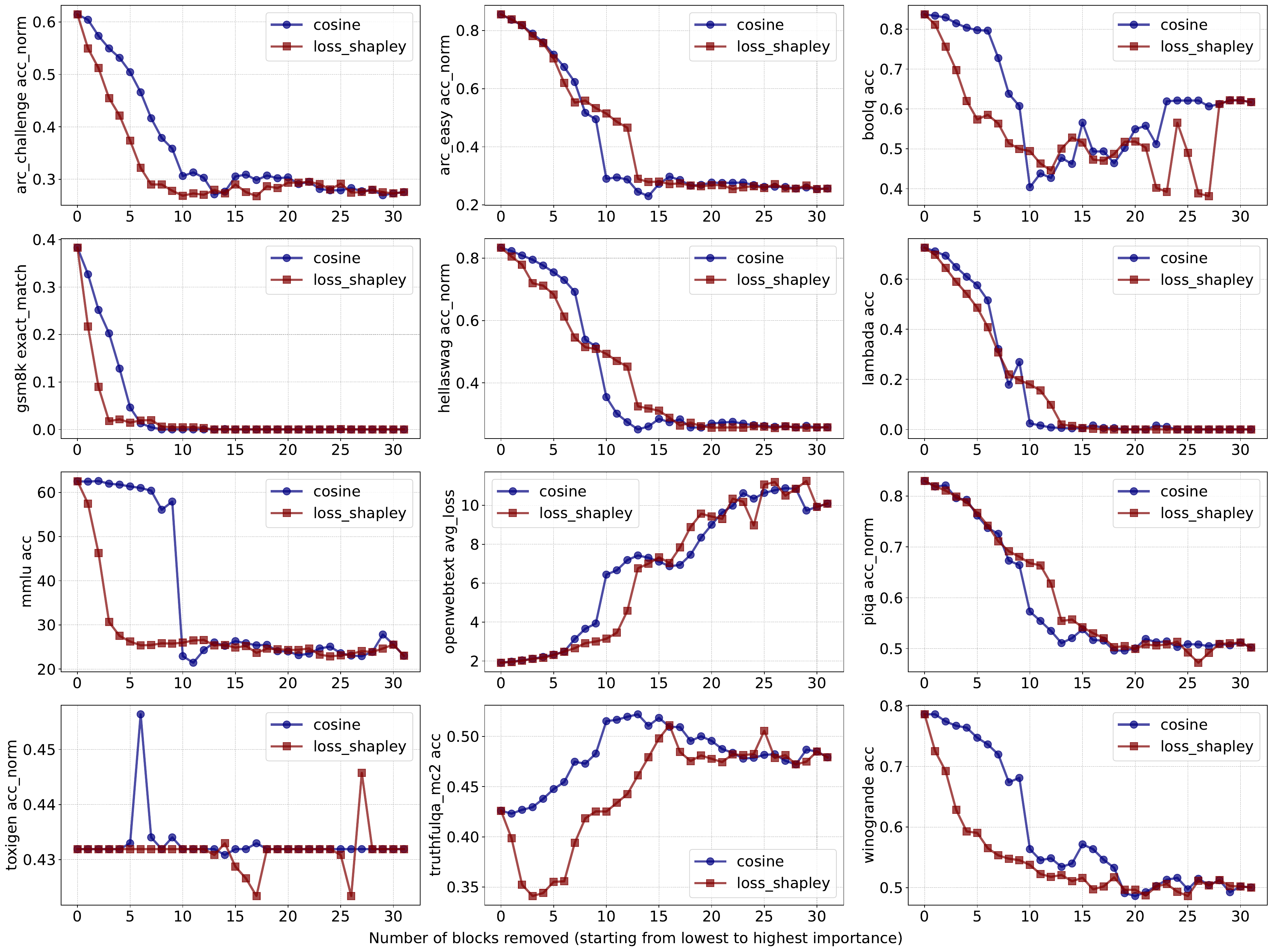}
    \caption{\textbf{Mistral-7b evaluation on multiple datasets from eval-harness~\citep{eval-harness}}, highlighting the impact of pruning on some tasks is more significant (such as GSM-8k~\citep{cobbe2021gsm8k}) as compared to others.}
    \label{fig:mistral_harness}
    \vspace{-5mm}
\end{figure}

We include tasks from OpenLLM leaderboard~\citep{open-llm-leaderboard} evaluated using lm-eval-harness package for reproducibility~\citep{eval-harness}.
In particular, we evaluate performance on MMLU~\citep{hendrycks2020mmlu}, GSM-8k~\citep{cobbe2021gsm8k}, ARC (easy as well as challenging)~\citep{clark2018arc}, BoolQ~\citep{clark2019boolq}, HellaSwag~\citep{zellers2019hellaswag}, Lambada~\citep{paperno2016lambada}, PiQA~\citep{bisk2020piqa}, Toxigen~\citep{hartvigsen2022toxigen}, TruthfulQA (MC2)~\citep{lin22truthfulqa}, and Winogrande~\citep{sakaguchi2019winogrande}.

Fig.~\ref{fig:llama2_harness} presents our results on LLaMa-2 7b~\citep{touvron2023llama2} while Fig.~\ref{fig:mistral_harness} presents the results on Mistral 7b~\citep{jiang2023mistral}.
We only plot the results for cosine block influence and loss shapley (computed on OpenWebText language modeling task).
The results highlight that the model suffers from performance degradation on some tasks such as GSM-8k~\citep{cobbe2021gsm8k} and ARC challenge~\citep{clark2018arc} even after pruning just a single block from the model.
This indicates that just looking at MMLU provides only a partial picture of the true impact of depth pruning on model performance.

\begin{figure*}[t]
    \centering
    \subfloat{
        \includegraphics[width=\textwidth]{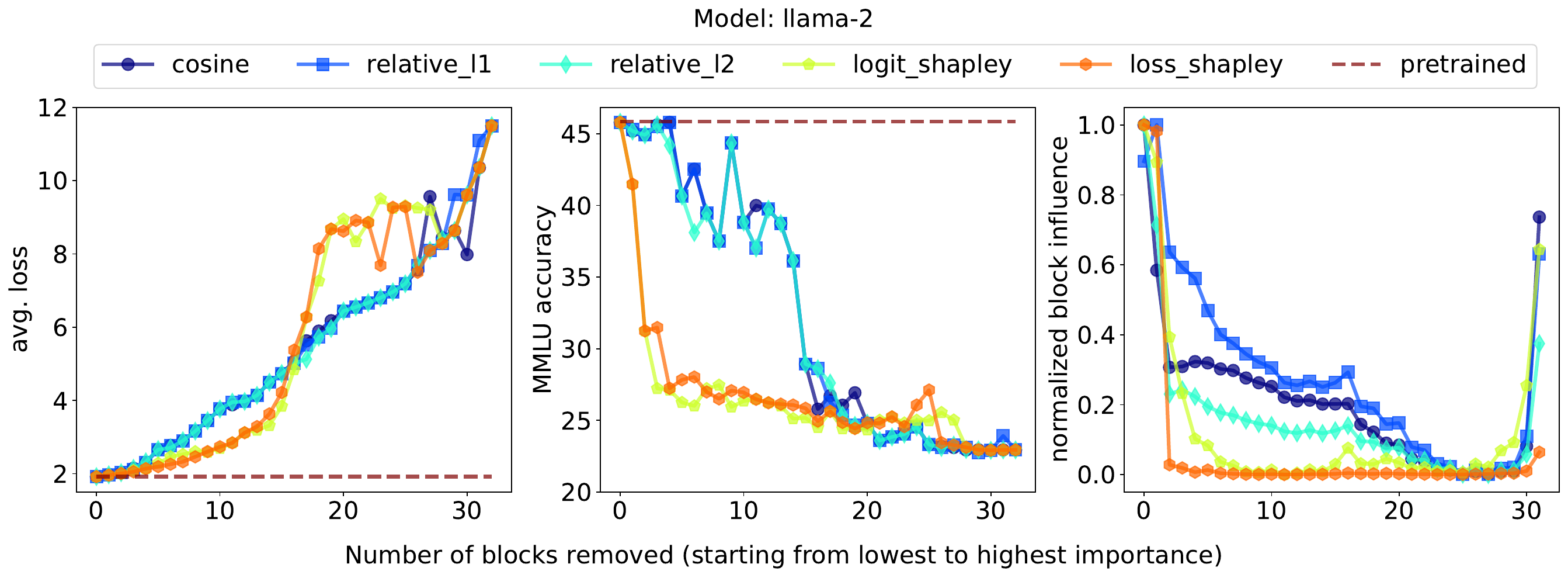}
    }

    \vspace{-3mm}
    \subfloat{
        \includegraphics[width=\textwidth]{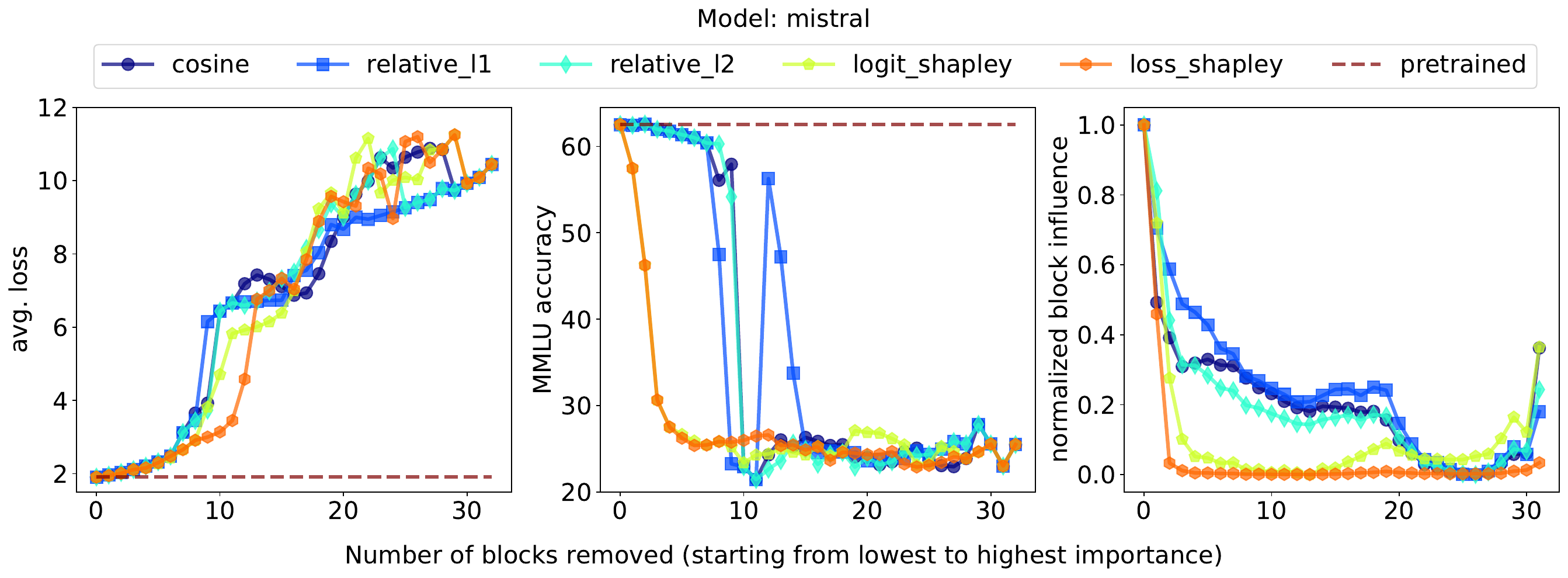}
    }
    \caption{\textbf{Comparison of different block influence metrics} (detailed in Section~\ref{subsec:block_inf_metrics}) used for block pruning and their impact on downstream performance on LLaMa-2 7b (top) and Mistral 7b (bottom) in terms of average loss on a small curated validation set from OpenWebText and accuracy on MMLU.}
    \label{fig:metrics_comparison}
    \vspace{-5mm}
\end{figure*}

\begin{figure}[t]
    \centering
    \subfloat{
        \includegraphics[width=\textwidth]{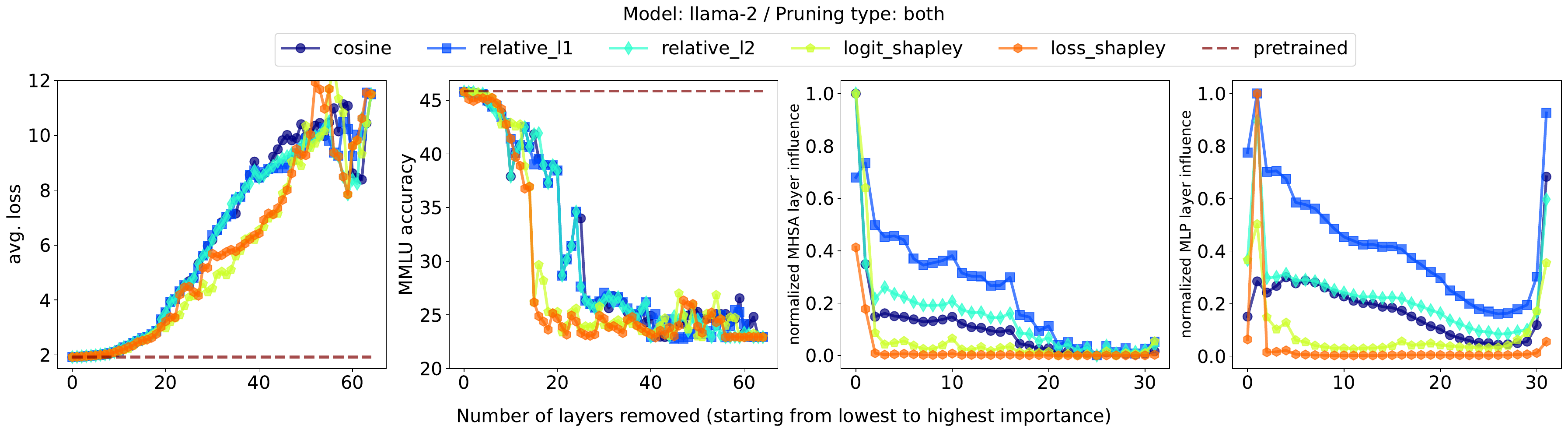}
    }

    \vspace{-3mm}
    \subfloat{
        \includegraphics[width=\textwidth]{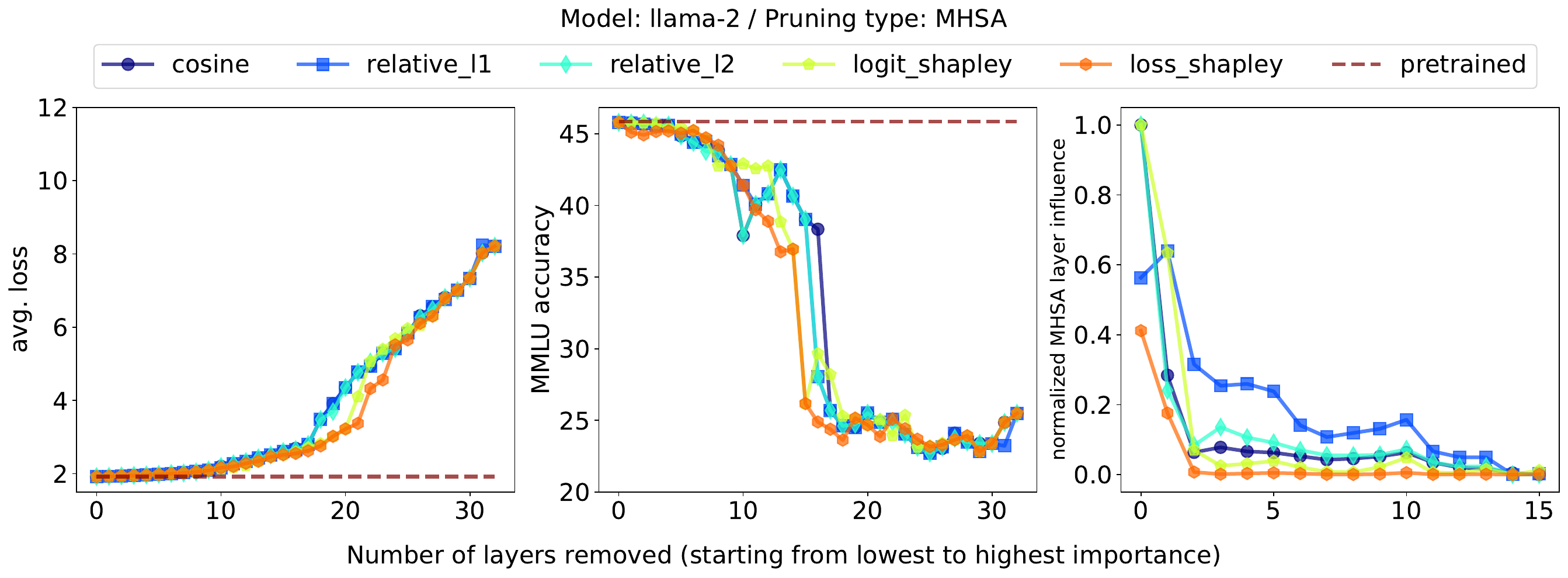}
    }

    \vspace{-3mm}
    \subfloat{
        \includegraphics[width=\textwidth]{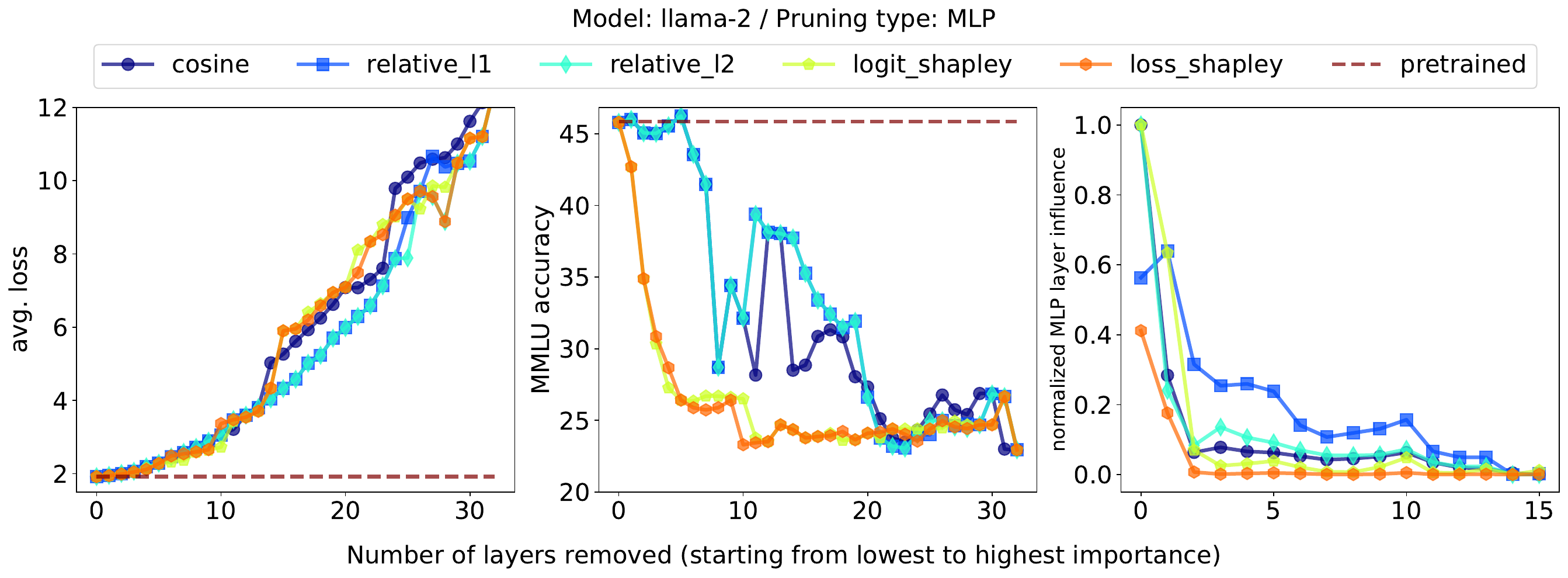}
    }
    \caption{\textbf{Comparison of different layer influence metrics on LLaMa-2 7b} (detailed in Section~\ref{subsec:block_inf_metrics}) and their impact on downstream performance in terms of average loss on a small curated validation set from OpenWebText and accuracy on MMLU. The first row highlights joint pruning, while the second and third row highlights pruning of just self-attention or feed-forward layer respectively.}
    \label{fig:layer_influence_llama2}
\end{figure}

\begin{figure*}[t]
    \centering
    \subfloat{
        \includegraphics[width=\textwidth]{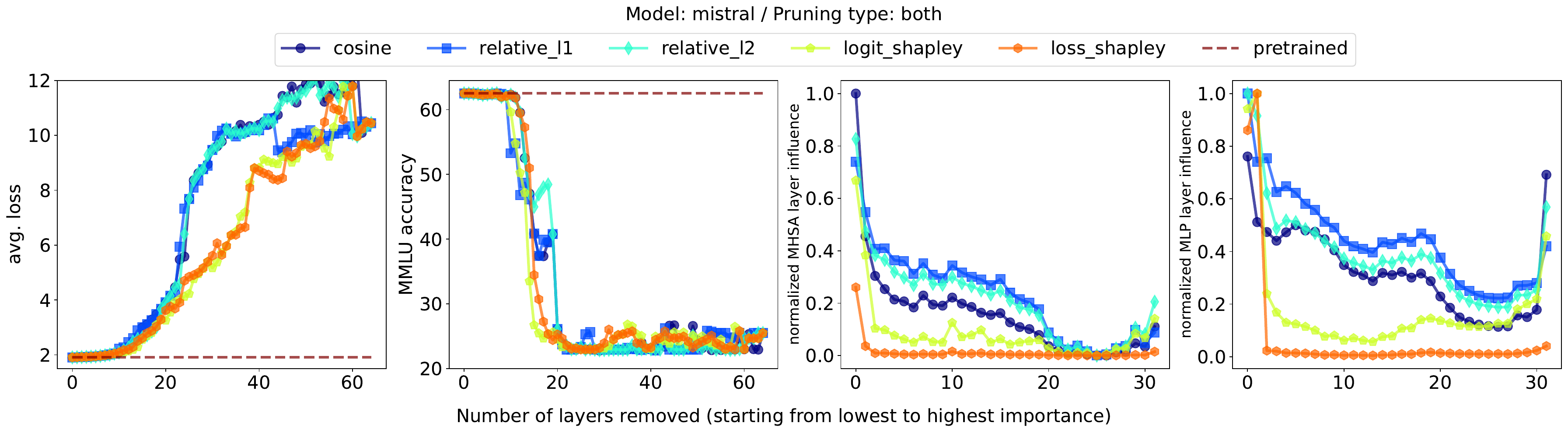}
    }

    \vspace{-3mm}
    \subfloat{
        \includegraphics[width=\textwidth]{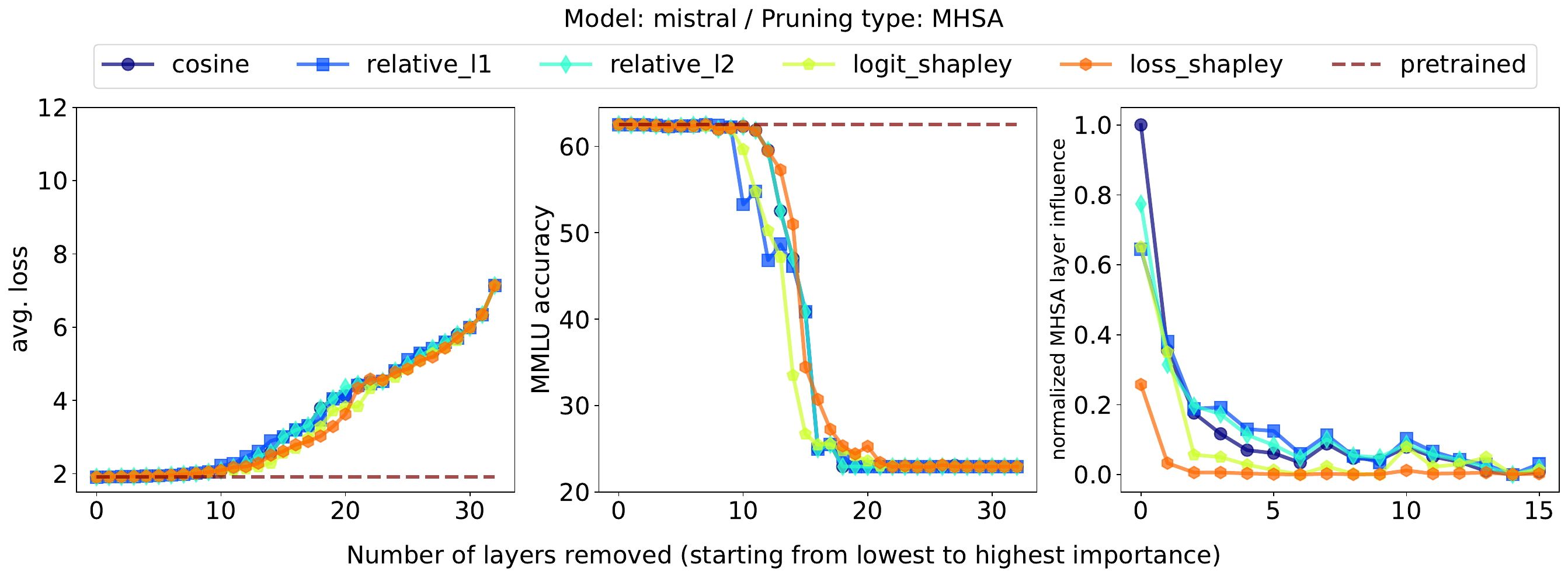}
    }

    \vspace{-3mm}
    \subfloat{
        \includegraphics[width=\textwidth]{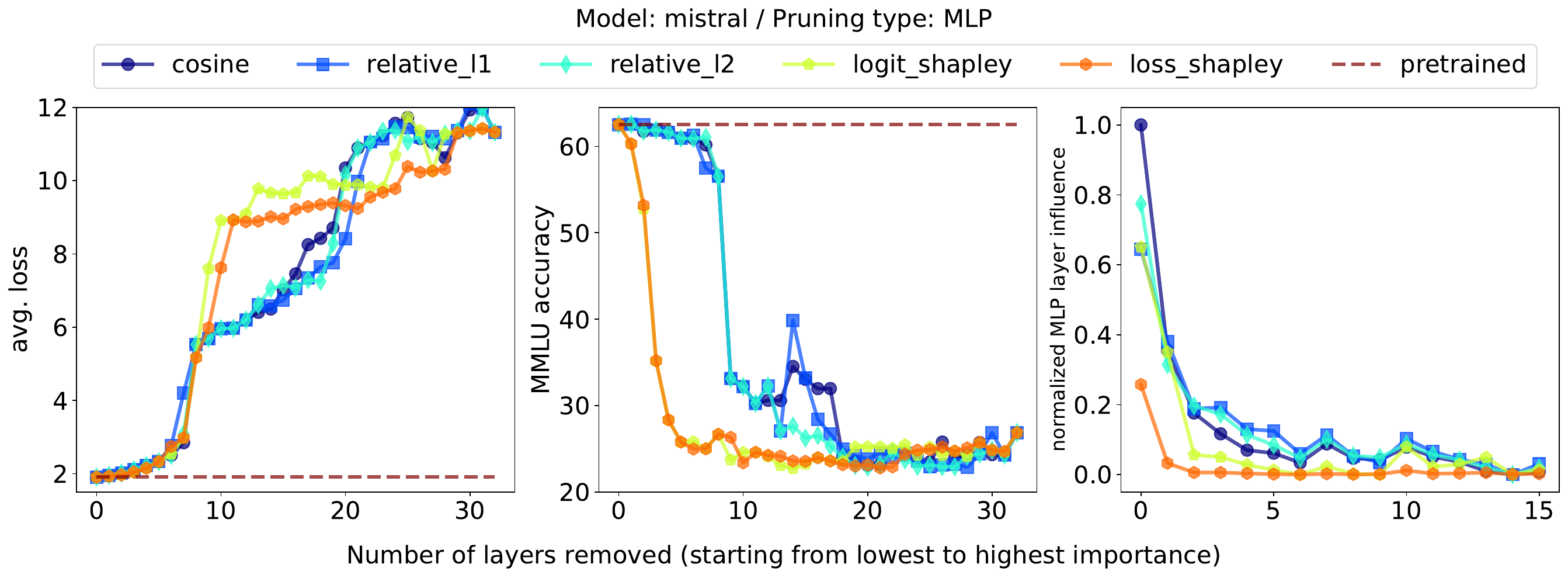}
    }
    \caption{\textbf{Comparison of different layer influence metrics on Mistral 7b} (detailed in Section~\ref{subsec:block_inf_metrics}) and their impact on downstream performance in terms of average loss on a small curated validation set from OpenWebText and accuracy on MMLU. The first row highlights joint pruning, while the second and third row highlights pruning of just self-attention or feed-forward layer respectively.} %\PM{quite interesting result! It seems that MHSA only pruning with Shapley should be better than cosine for full blocks, right? Comparing Figure 1 and Figure 5 it seems that you already have improvements by pruning both layers independently. }}
    \vspace{-5mm}
    \label{fig:layer_influence_mistral}
\end{figure*}

\begin{figure*}[t]
    \centering
    \subfloat{
        \includegraphics[width=\textwidth]{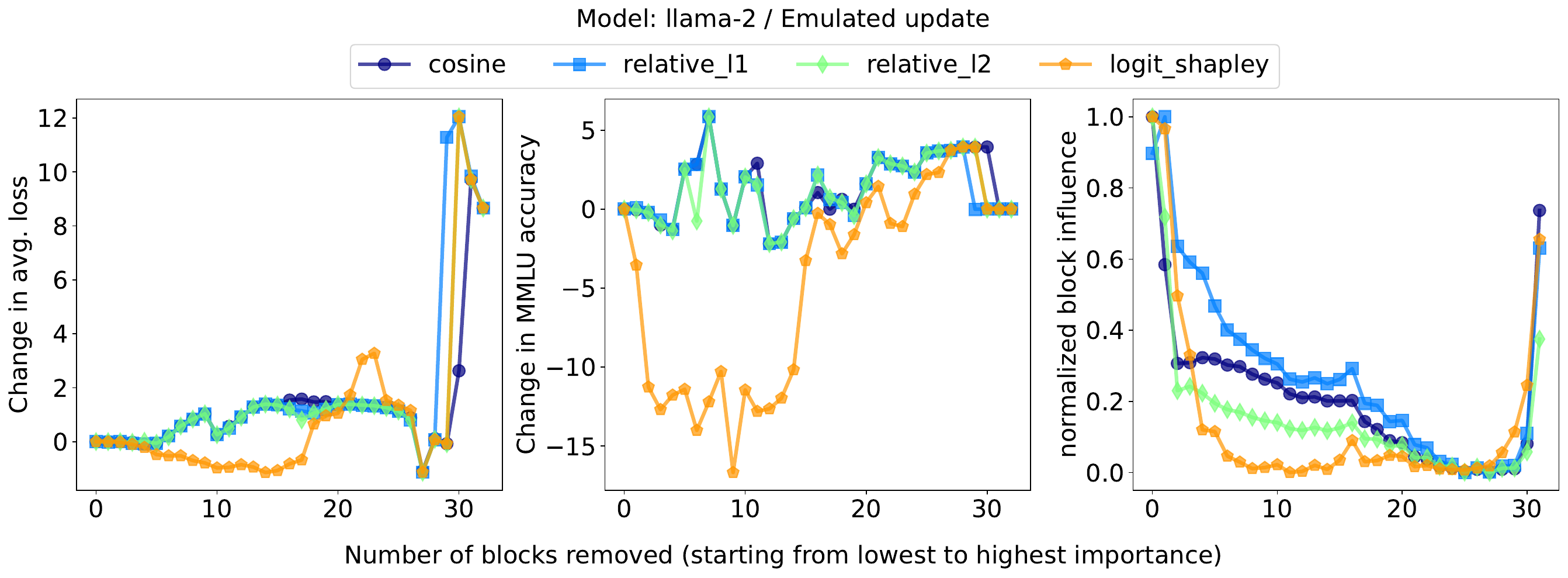}
    }

    \vspace{-3mm}
    \subfloat{
        \includegraphics[width=\textwidth]{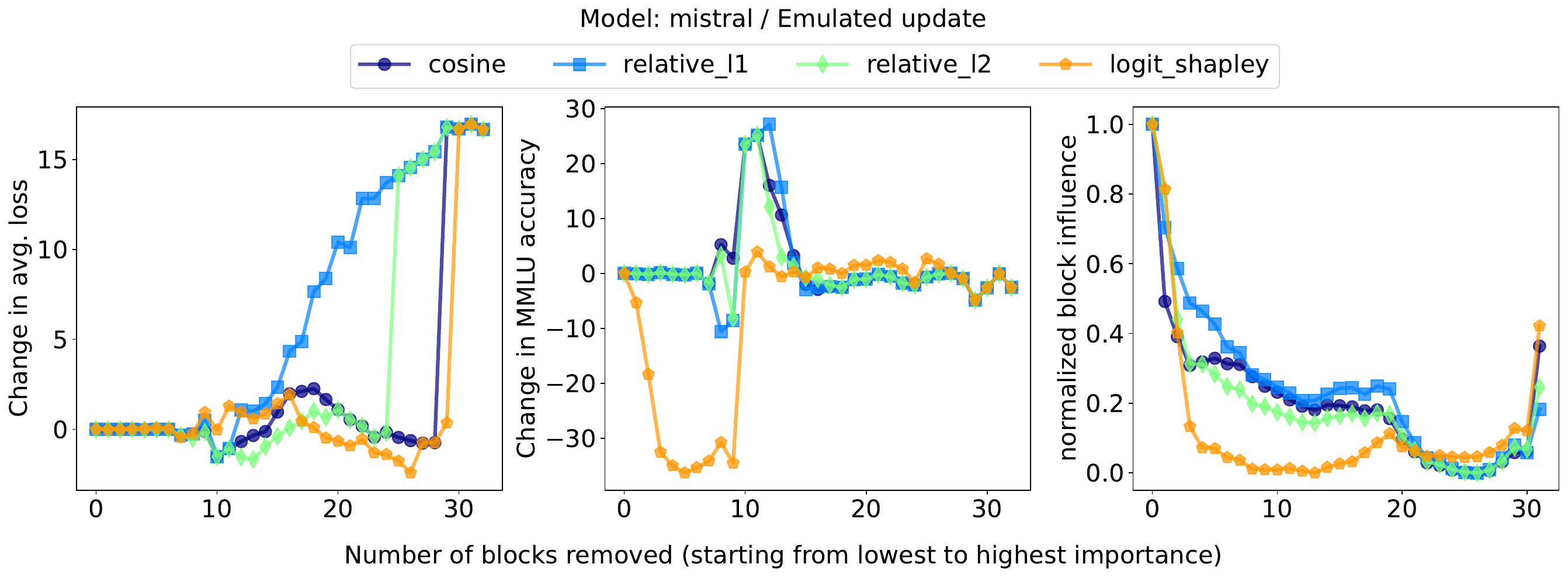}
    }
    \caption{\textbf{Relative impact on performance with emulated update as a performance recovery method.} The plot with the original values is presented in Fig.~\ref{fig:emulated_update}.}
    \label{fig:emulated_update_relative}
    \vspace{-5mm}
\end{figure*}

\begin{figure}
    \centering
    \subfloat{
        \includegraphics[width=\textwidth]{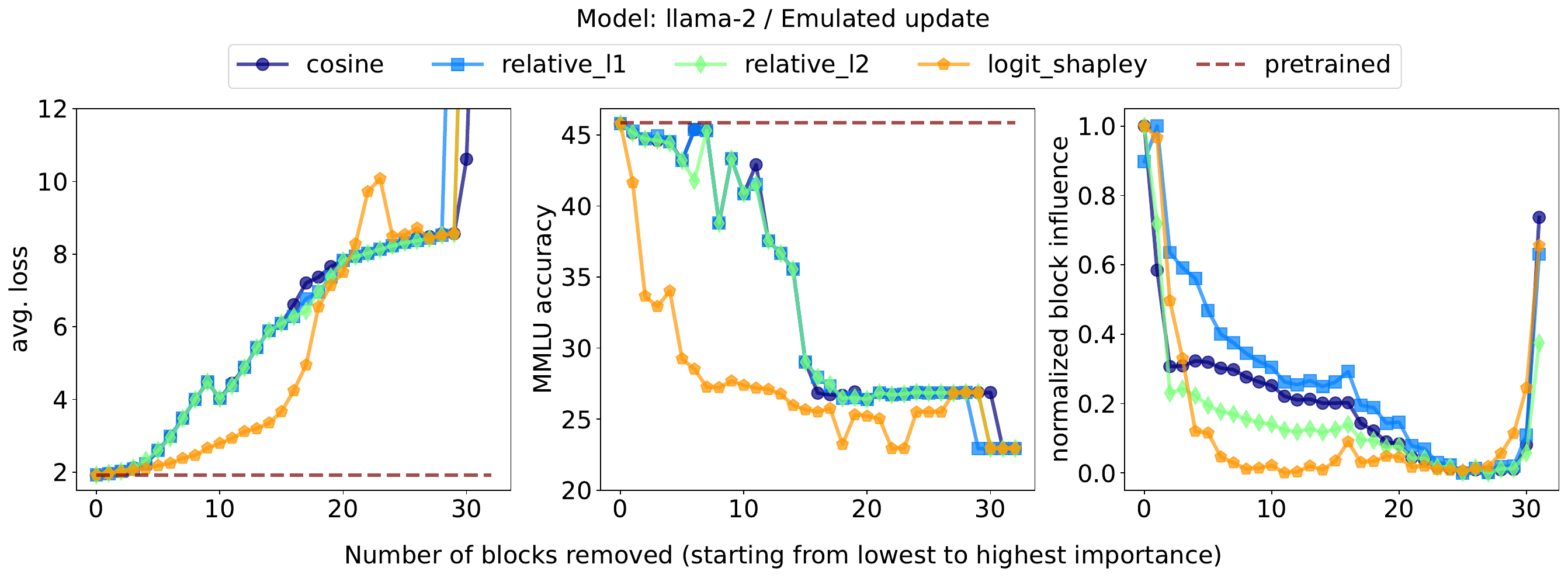}
    }
    
    \subfloat{
        \includegraphics[width=\textwidth]{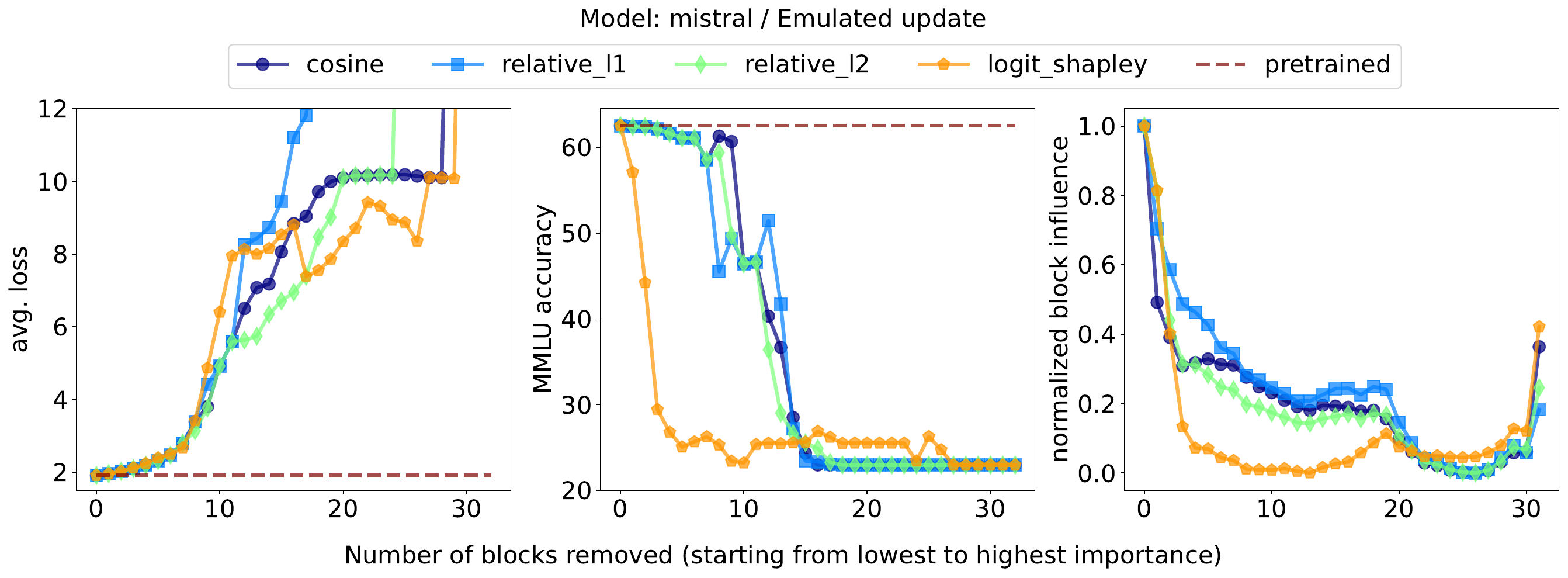}
    }
    \caption{\textbf{Impact on performance with emulated update as a performance recovery method.}}
    \label{fig:emulated_update}
\end{figure}

\section{Low-rank linear adapters.}
\label{appendix:low_rank_results}

\begin{figure}
    \centering
    \subfloat{
        \includegraphics[width=\textwidth]{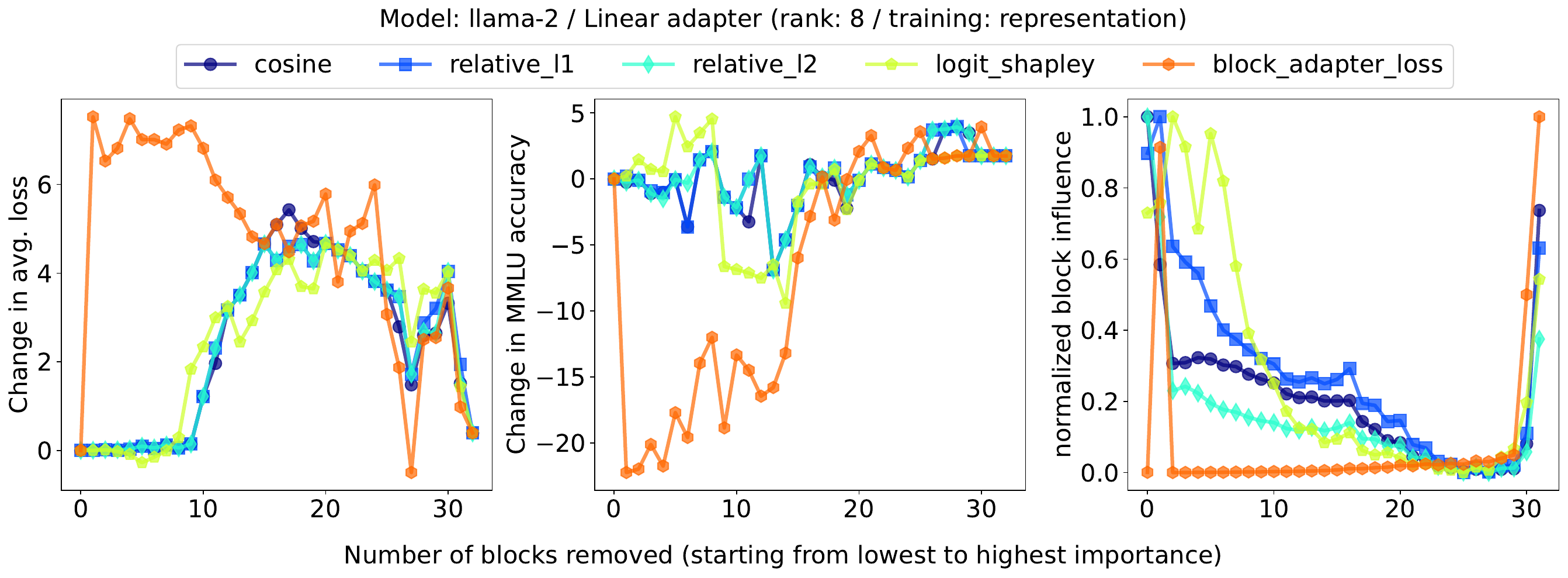}
    }

    \vspace{-3mm}
    \subfloat{
        \includegraphics[width=\textwidth]{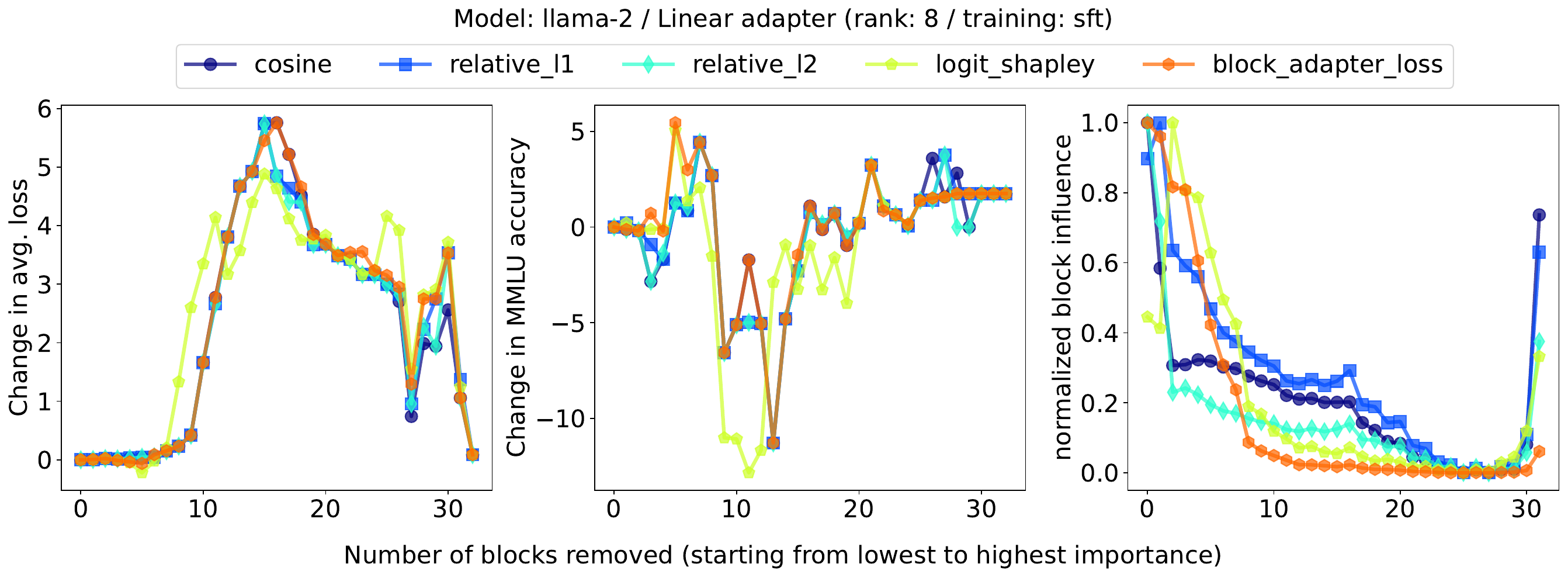}
    }

    \vspace{-3mm}
    \subfloat{
        \includegraphics[width=\textwidth]{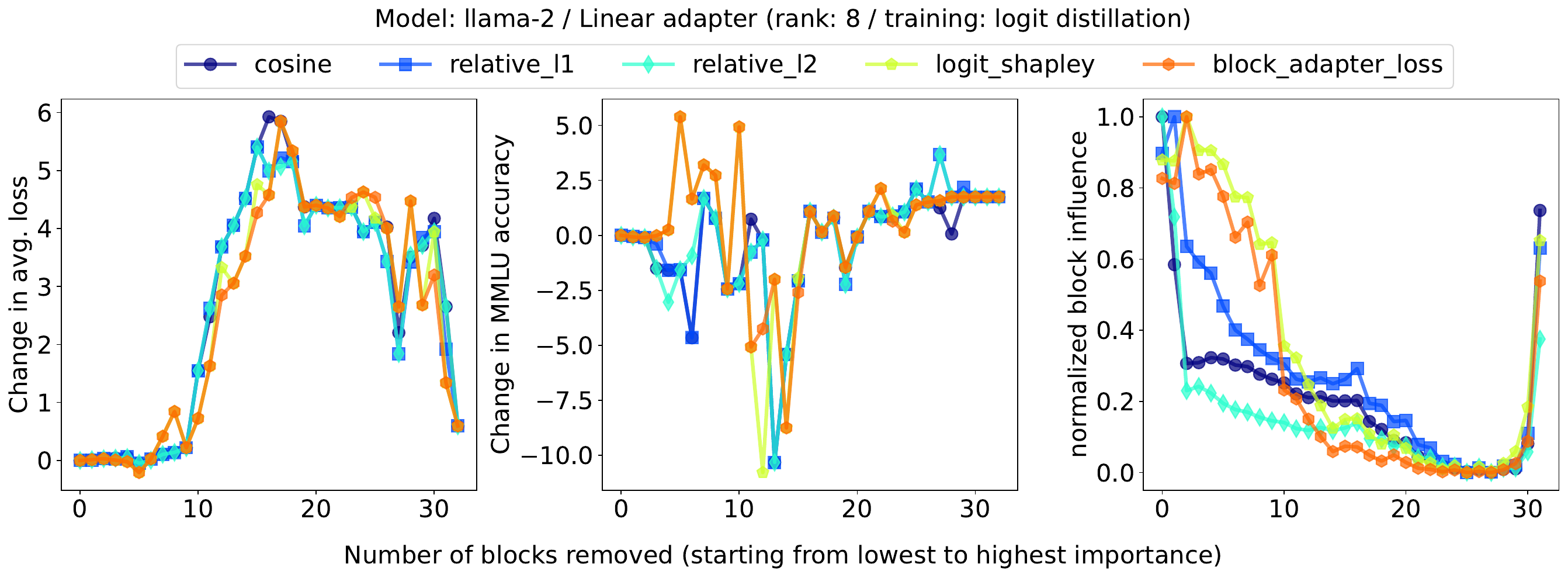}
    }
    \caption{\textbf{Evaluating the relative impact of linear adapters for LLaMa-2 7b with a rank of 8} trained using three different metrics including (a) MSE loss defined on the representation, (b) supervised fine-tuning (SFT), and (c) logit distillation where logits are distilled from the full model. The plot with the original values is presented in Fig.~\ref{fig:linear_adapters_rank_8_llama2}.}
    \label{fig:linear_adapters_rank_8_llama2_relative}
\end{figure}

\begin{figure*}[t]
    \centering
    \subfloat{
        \includegraphics[width=\textwidth]{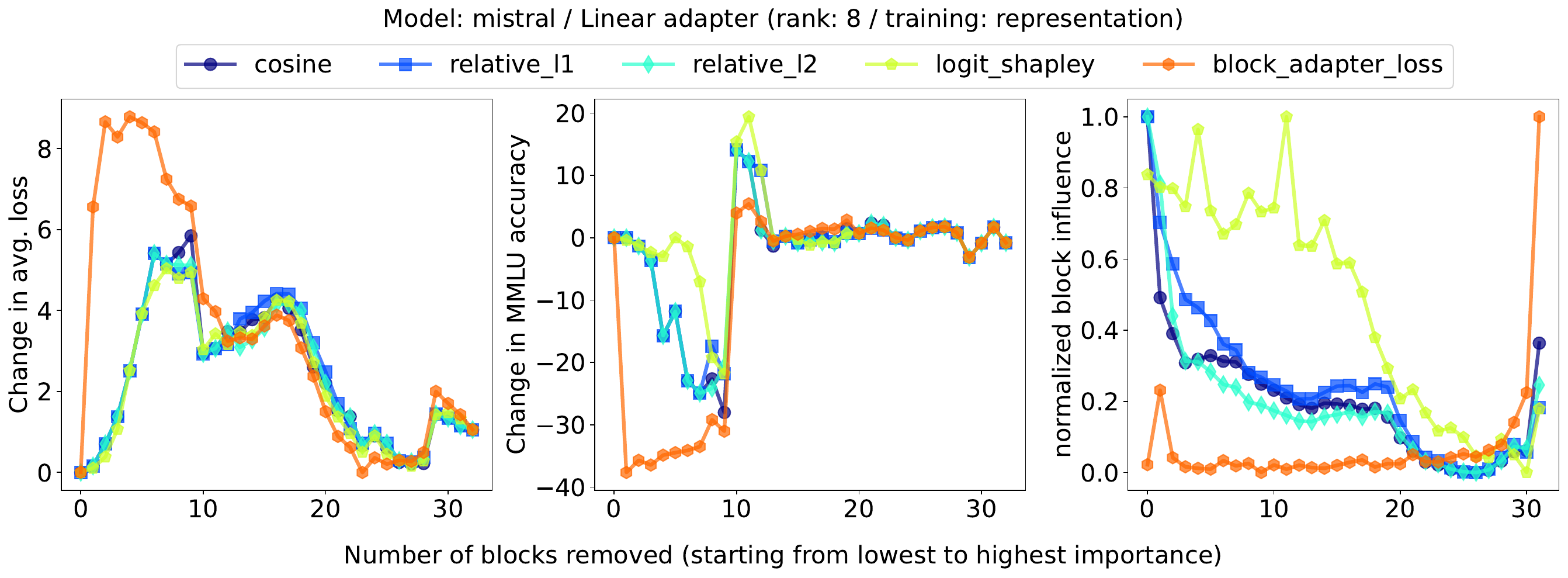}
    }

    \vspace{-3mm}
    \subfloat{
        \includegraphics[width=\textwidth]{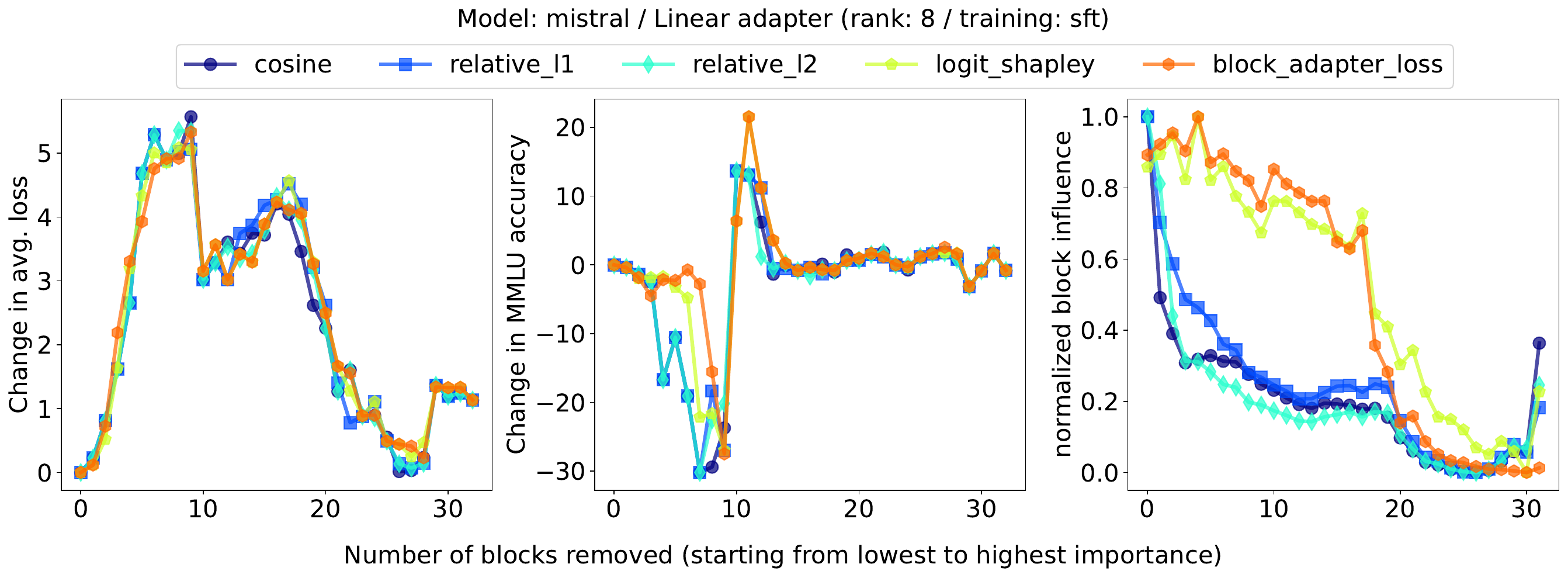}
    }

    \vspace{-3mm}
    \subfloat{
        \includegraphics[width=\textwidth]{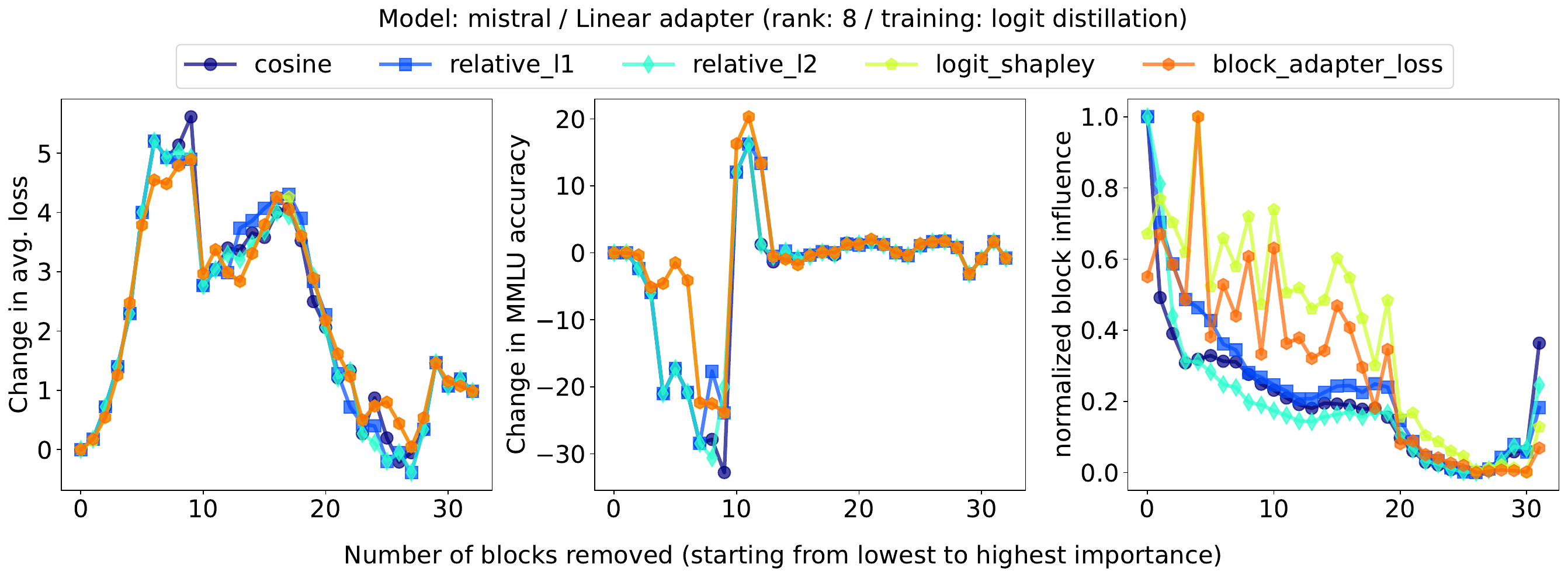}
    }
    \caption{\textbf{Evaluating the relative impact of linear adapters for Mistral 7b with a rank of 8} trained using three different metrics including (a) MSE loss defined on the representation, (b) supervised fine-tuning (SFT), and (c) logit distillation where logits are distilled from the full model. The plot with the original values is presented in Fig.~\ref{fig:linear_adapters_rank_8_mistral}.}
    \label{fig:linear_adapters_rank_8_mistral_relative}
    \vspace{-5mm}
\end{figure*}

\begin{figure}
    \centering
    \subfloat{
        \includegraphics[width=\textwidth]{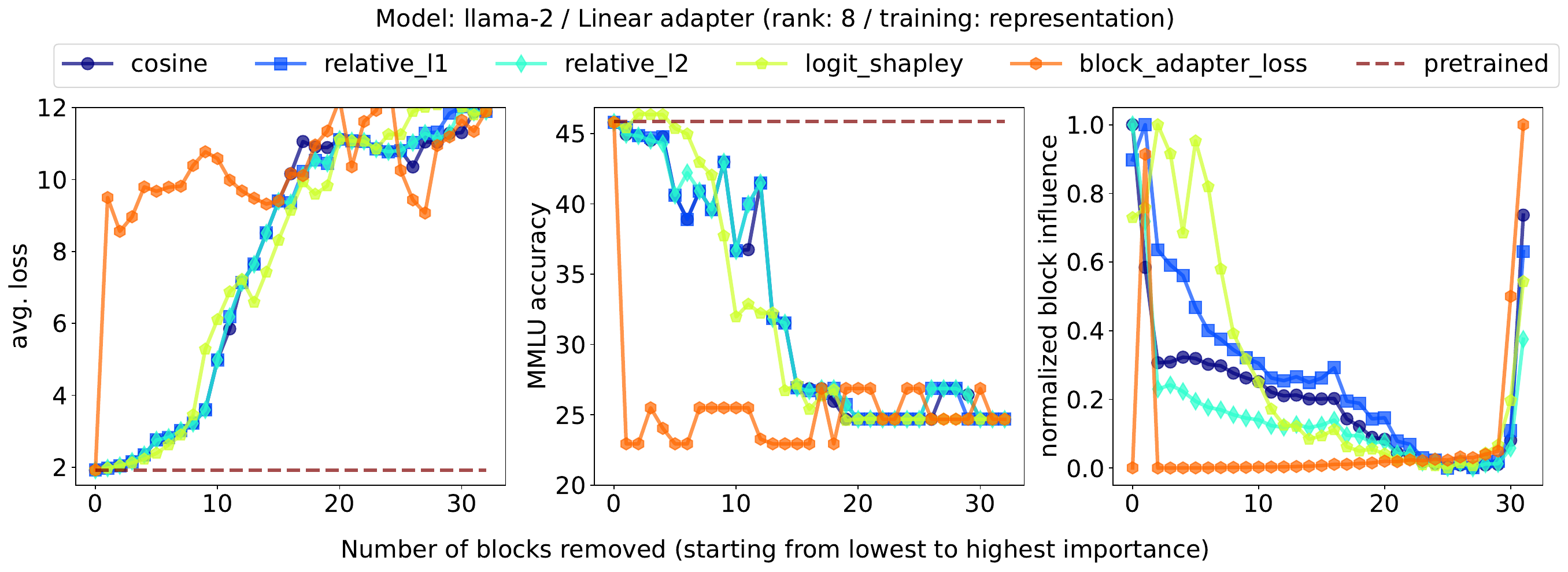}
    }

    \subfloat{
        \includegraphics[width=\textwidth]{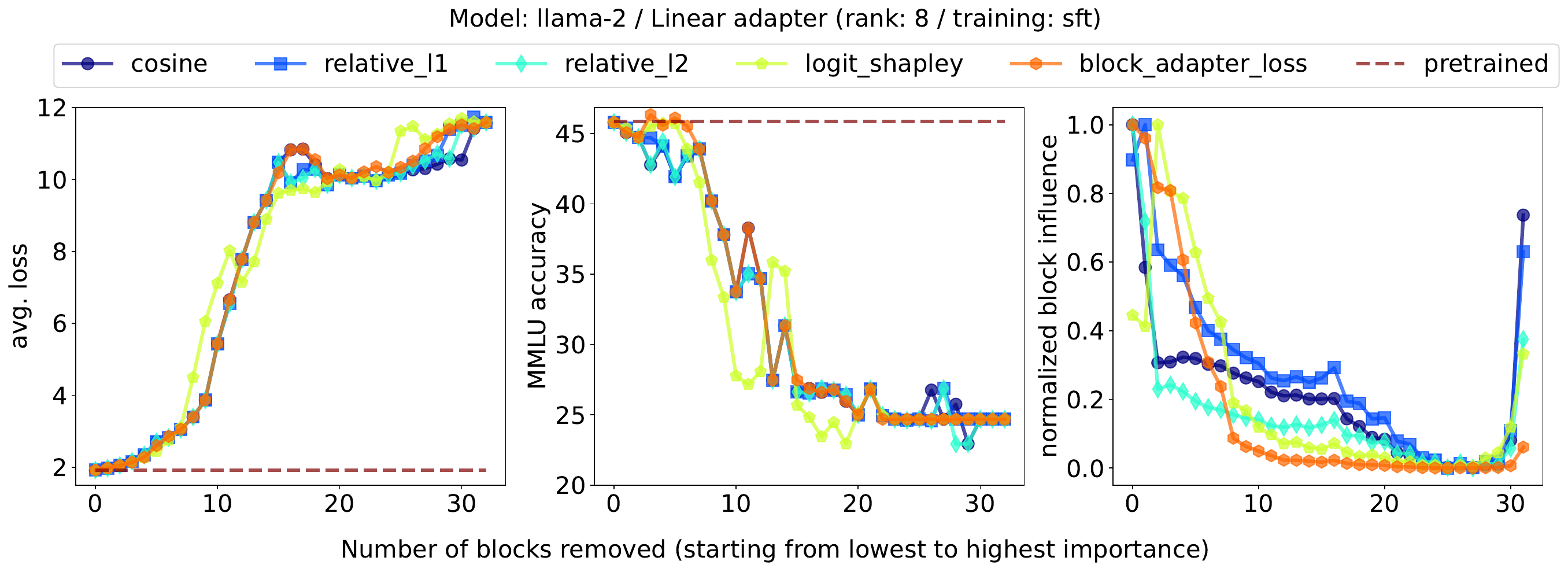}
    }

    \subfloat{
        \includegraphics[width=\textwidth]{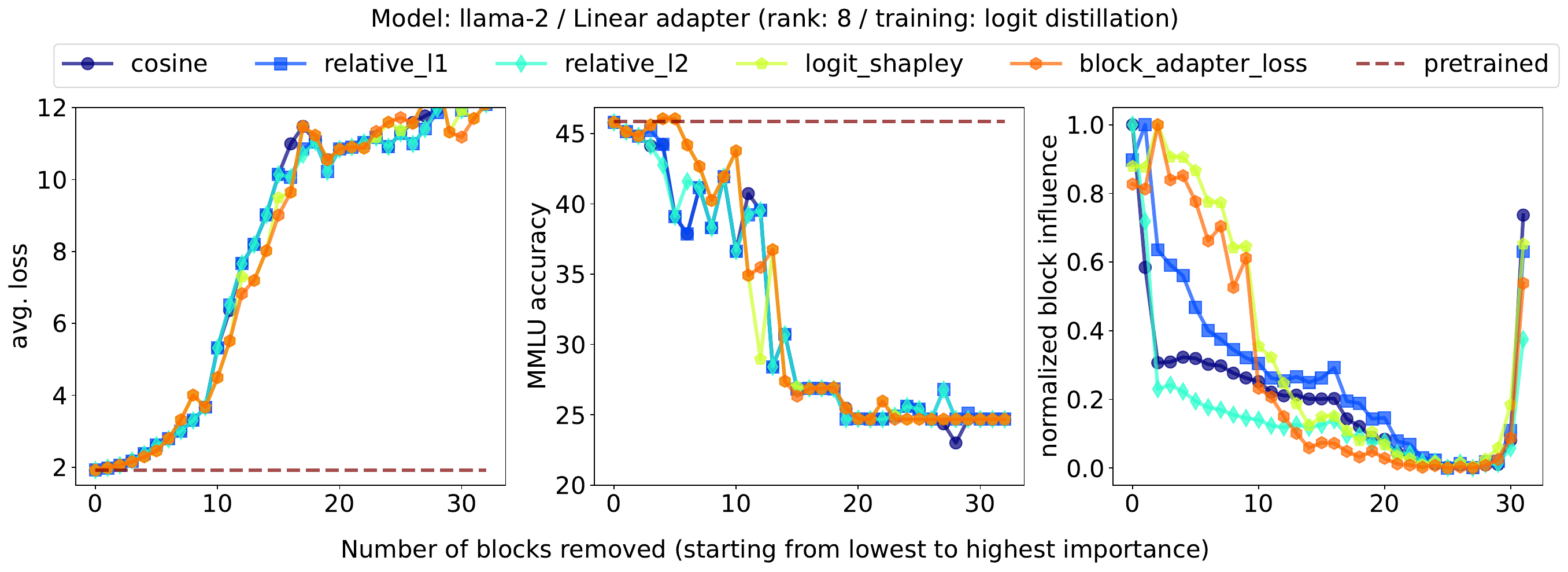}
    }
    \caption{\textbf{Evaluating the impact of linear adapters for LLaMa-2 7b with a rank of 8} trained using three different metrics including (a) MSE loss defined on the representation, (b) supervised fine-tuning (SFT), and (c) logit distillation where logits are distilled from the full model.}
    \label{fig:linear_adapters_rank_8_llama2}
\end{figure}

\begin{figure}
    \centering
    \subfloat{
        \includegraphics[width=\textwidth]{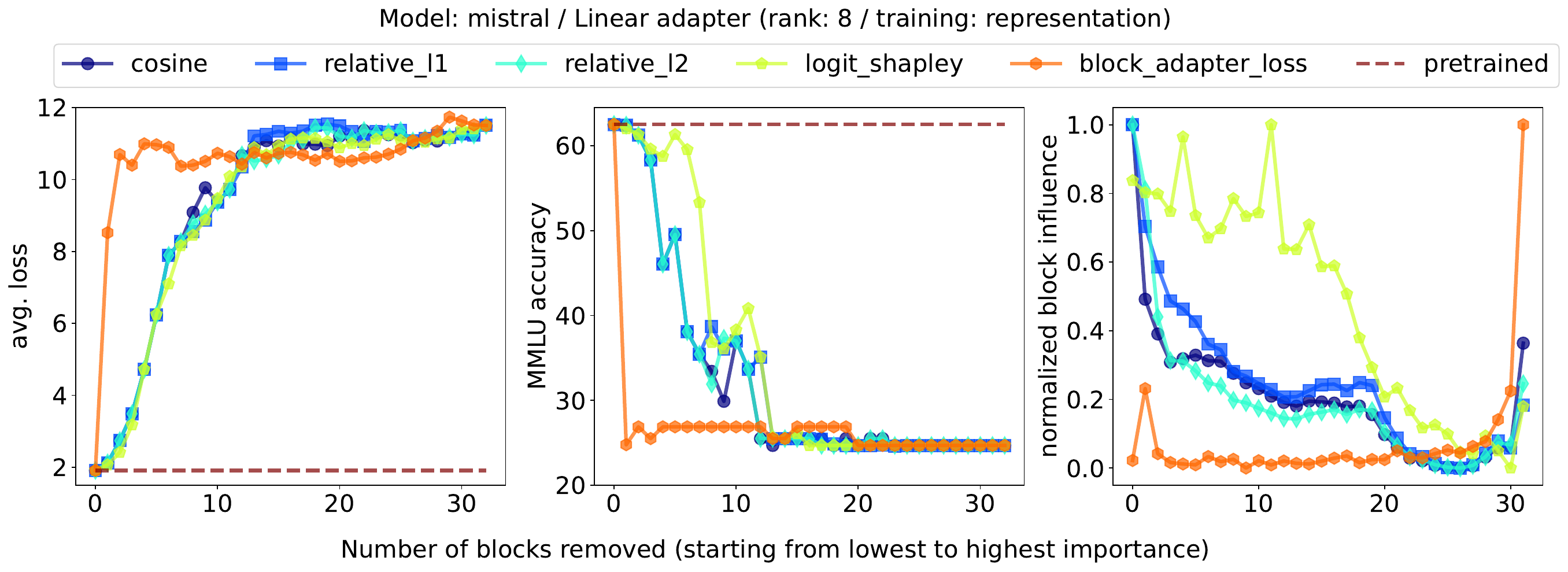}
    }

    \subfloat{
        \includegraphics[width=\textwidth]{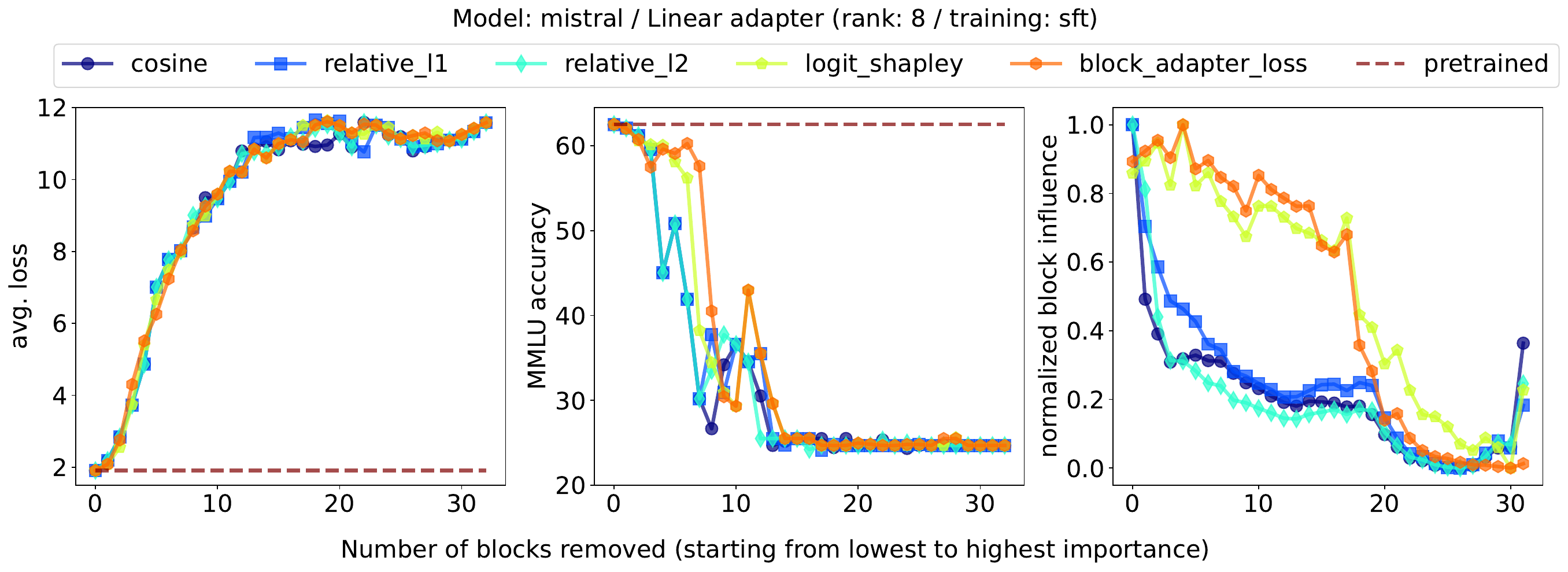}
    }

    \subfloat{
        \includegraphics[width=\textwidth]{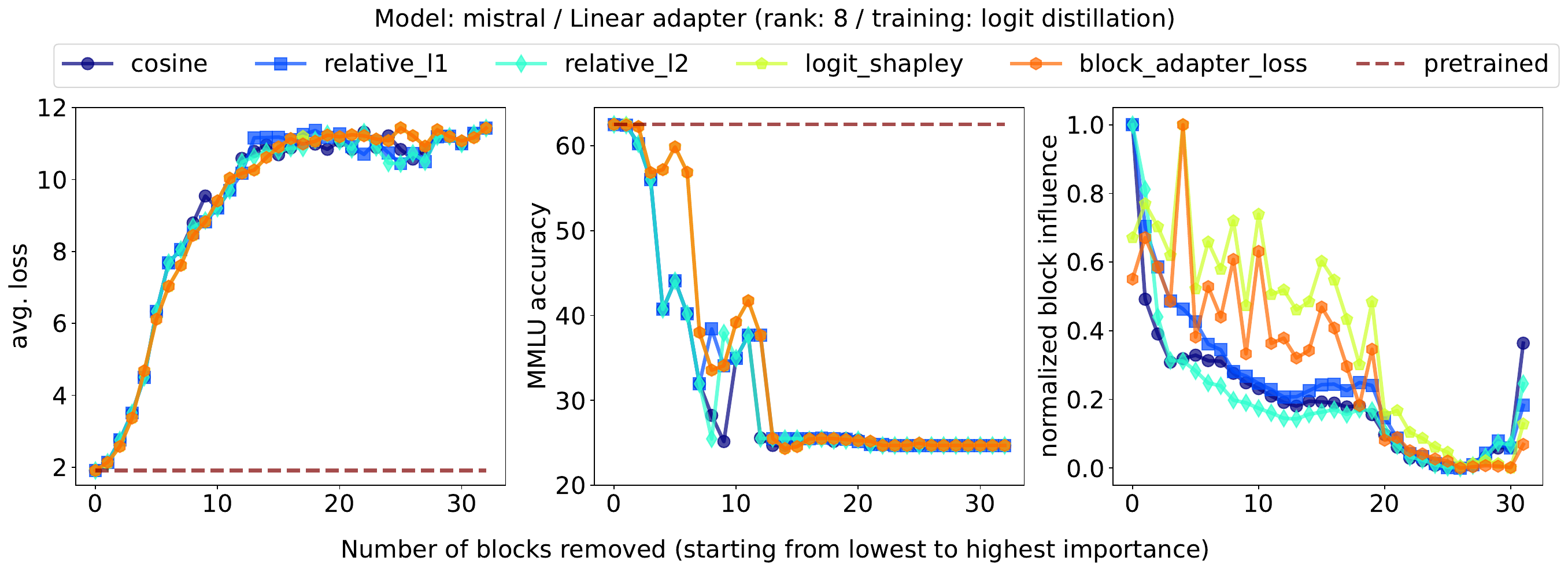}
    }
    \caption{\textbf{Evaluating the impact of linear adapters for Mistral 7b with a rank of 8} trained using three different metrics including (a) MSE loss defined on the representation, (b) supervised fine-tuning (SFT), and (c) logit distillation where logits are distilled from the full model.}
    \label{fig:linear_adapters_rank_8_mistral}
\end{figure}

Results for LLaMa-2 7b and Mistral 7b with an adapter rank of 8 on a relative scale are visualized in Fig.~\ref{fig:linear_adapters_rank_8_llama2_relative} and
Fig.~\ref{fig:linear_adapters_rank_8_mistral_relative} respectively.
Results for LLaMa-2 7b and Mistral 7b with an adapter rank of 8 on the original scale are visualized in 
Fig.~\ref{fig:linear_adapters_rank_8_llama2} and
Fig.~\ref{fig:linear_adapters_rank_8_mistral} respectively.

\begin{figure}
    \centering
    \subfloat{
        \includegraphics[width=\textwidth]{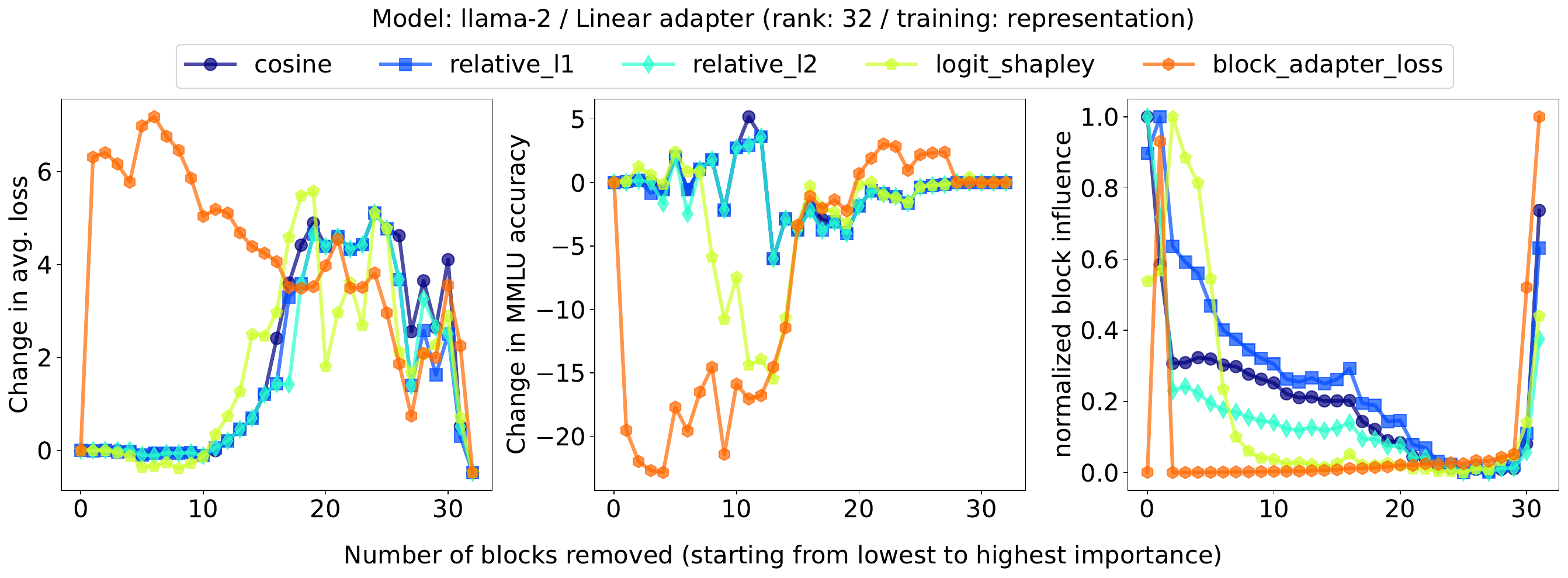}
    }

    \subfloat{
        \includegraphics[width=\textwidth]{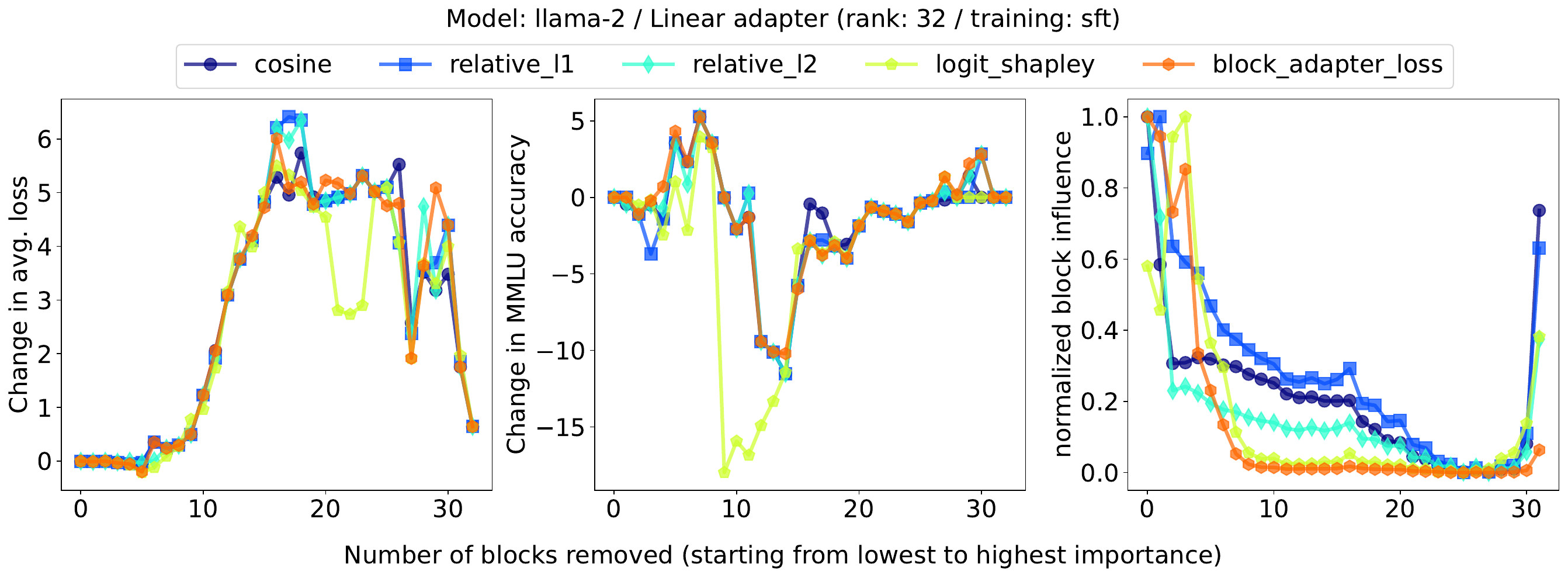}
    }

    \subfloat{
        \includegraphics[width=\textwidth]{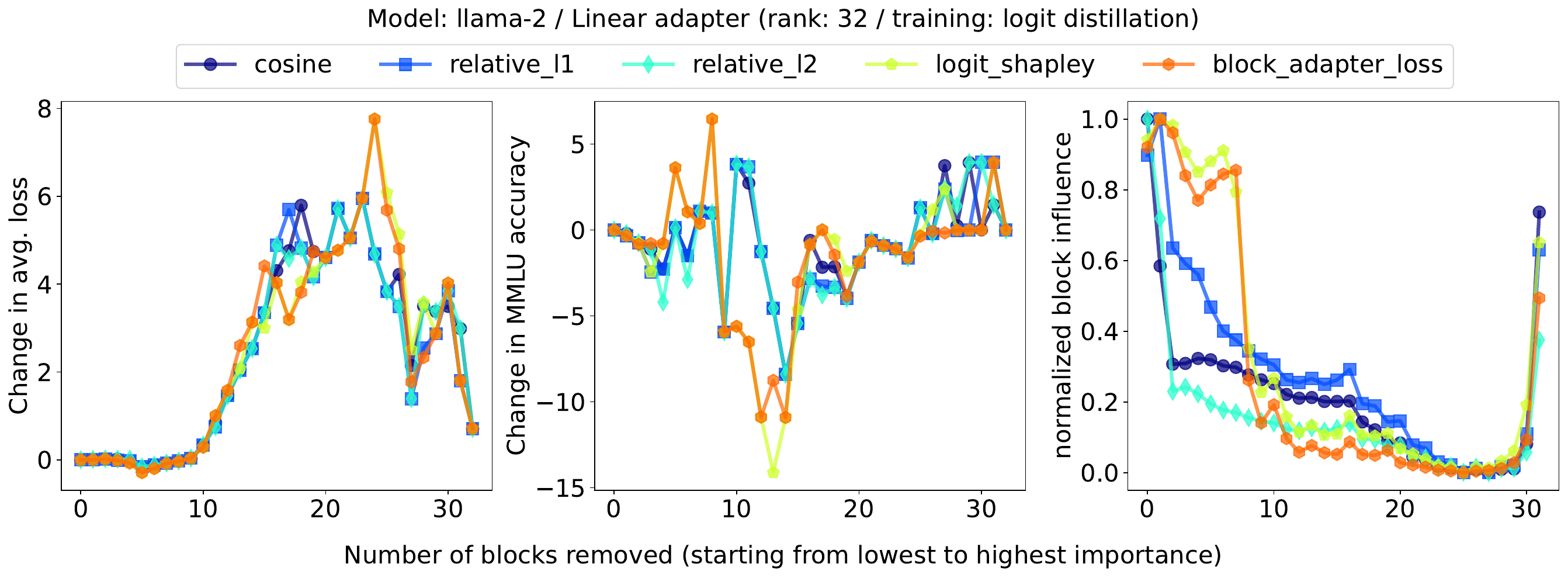}
    }
    \caption{\textbf{Evaluating the relative impact of linear adapters for LLaMa-2 7b with a rank of 32} trained using three different metrics including (a) MSE loss defined on the representation, (b) supervised fine-tuning (SFT), and (c) logit distillation where logits are distilled from the full model. The plot with the original values is presented in Fig.~\ref{fig:linear_adapters_rank_32_llama2}.}
    \label{fig:linear_adapters_rank_32_llama2_relative}
\end{figure}

\begin{figure}
    \centering
    \subfloat{
        \includegraphics[width=\textwidth]{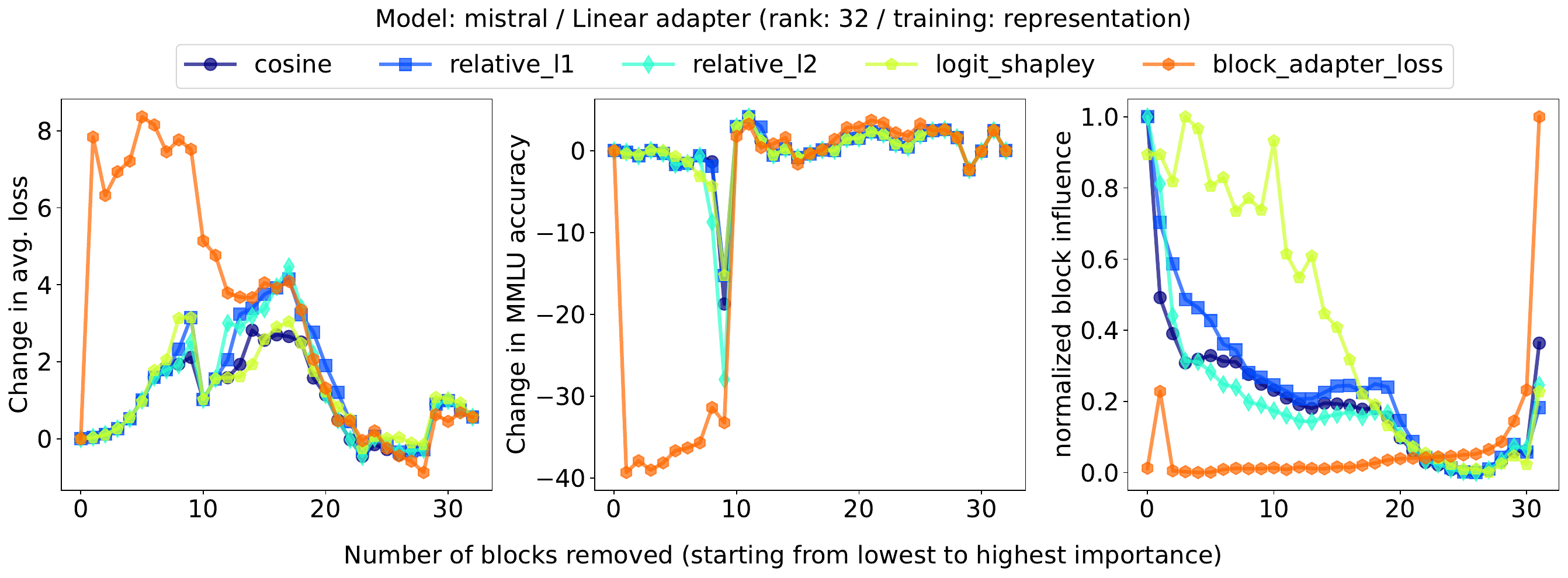}
    }

    \subfloat{
        \includegraphics[width=\textwidth]{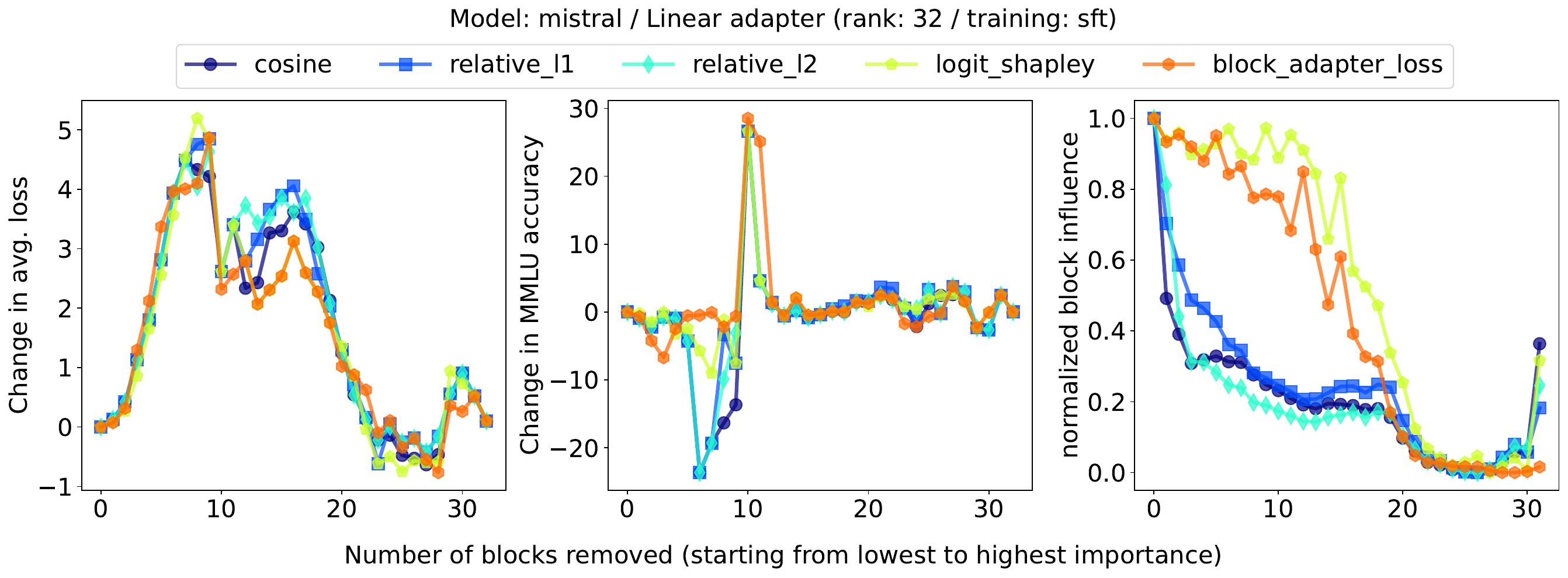}
    }

    \subfloat{
        \includegraphics[width=\textwidth]{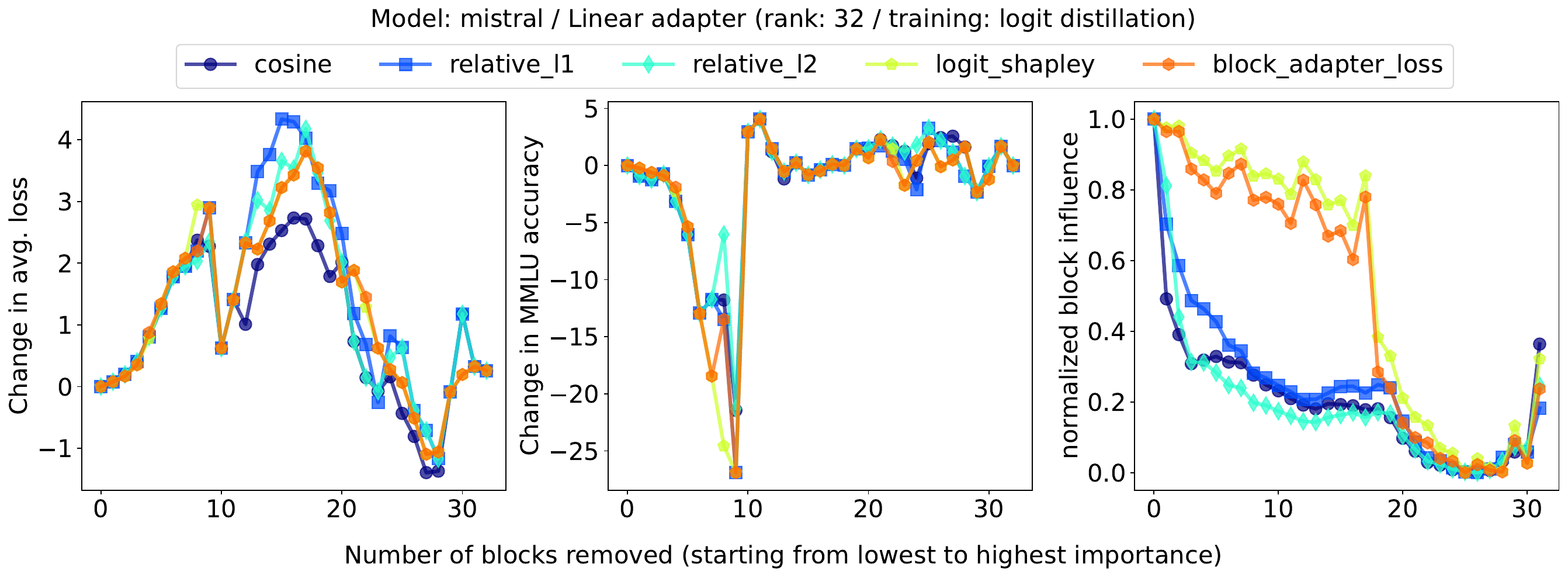}
    }
    \caption{\textbf{Evaluating the relative impact of linear adapters for Mistral 7b with a rank of 32} trained using three different metrics including (a) MSE loss defined on the representation, (b) supervised fine-tuning (SFT), and (c) logit distillation where logits are distilled from the full model. The plot with the original values is presented in Fig.~\ref{fig:linear_adapters_rank_32_mistral}.}
    \label{fig:linear_adapters_rank_32_mistral_relative}
\end{figure}

\begin{figure}
    \centering
    \subfloat{
        \includegraphics[width=\textwidth]{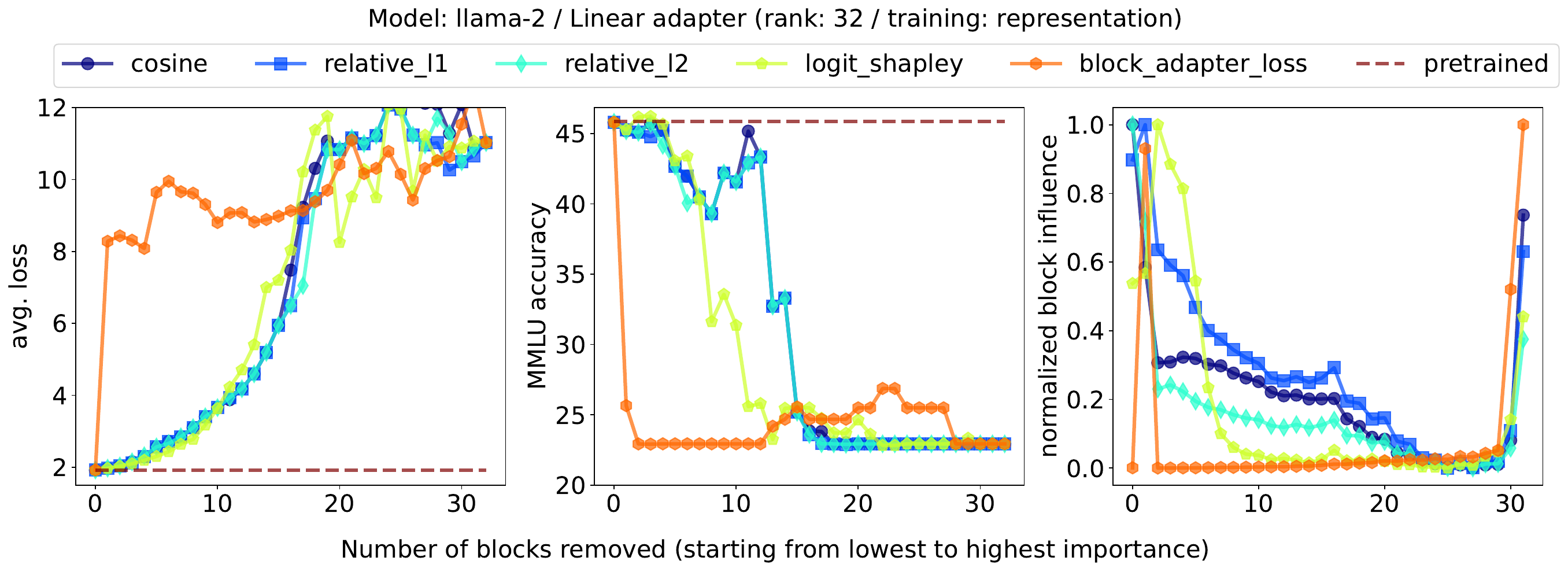}
    }

    \subfloat{
        \includegraphics[width=\textwidth]{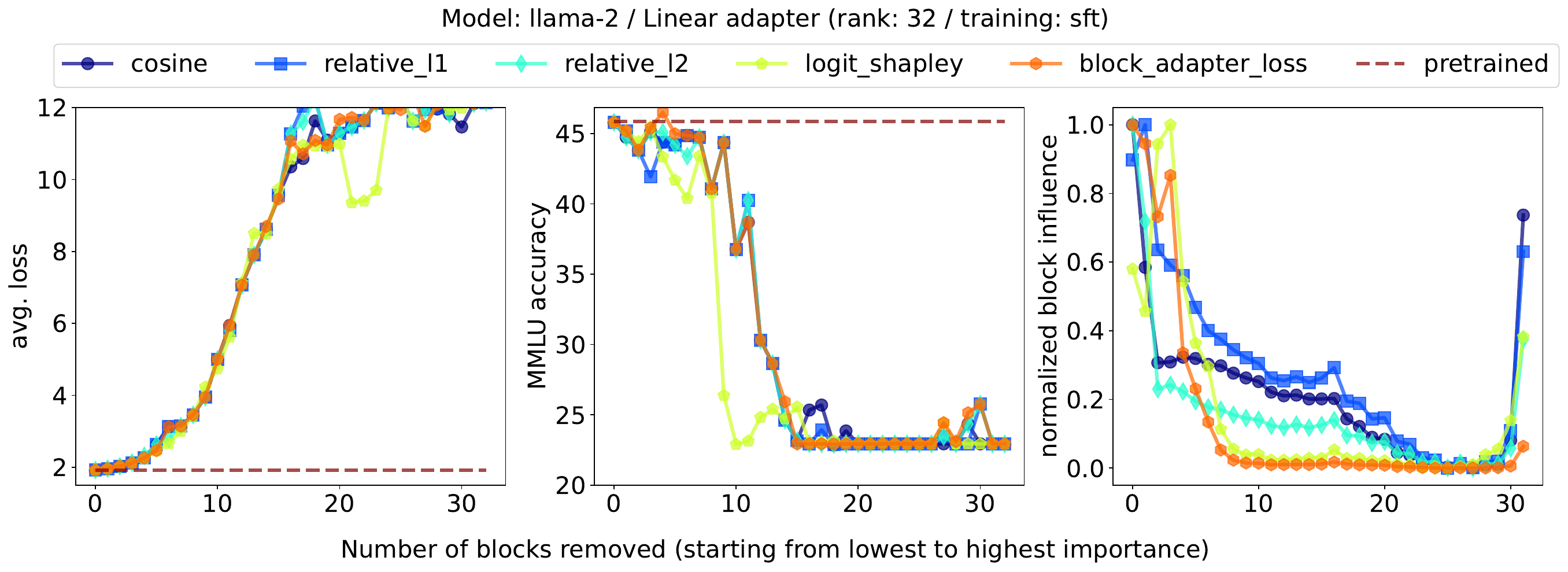}
    }

    \subfloat{
        \includegraphics[width=\textwidth]{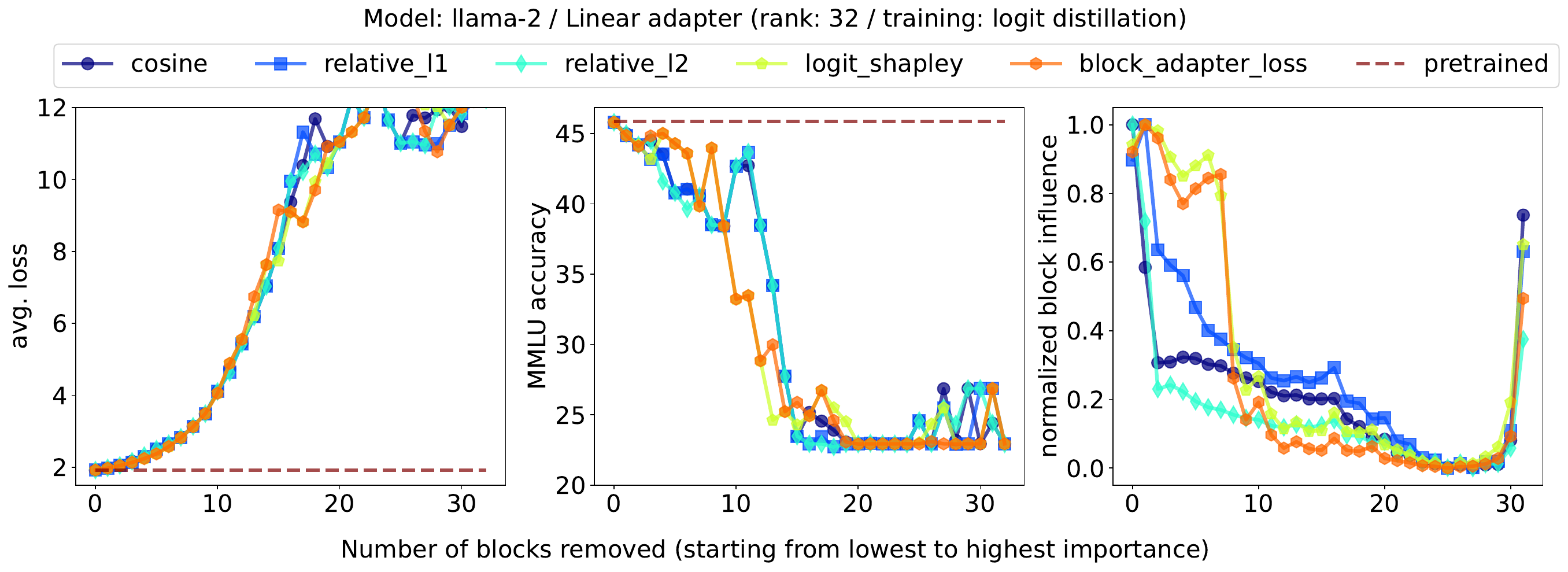}
    }
    \caption{\textbf{Evaluating the impact of linear adapters for LLaMa-2 7b with a rank of 32} trained using three different metrics including (a) MSE loss defined on the representation, (b) supervised fine-tuning (SFT), and (c) logit distillation where logits are distilled from the full model.}
    \label{fig:linear_adapters_rank_32_llama2}
\end{figure}

\begin{figure}
    \centering
    \subfloat{
        \includegraphics[width=\textwidth]{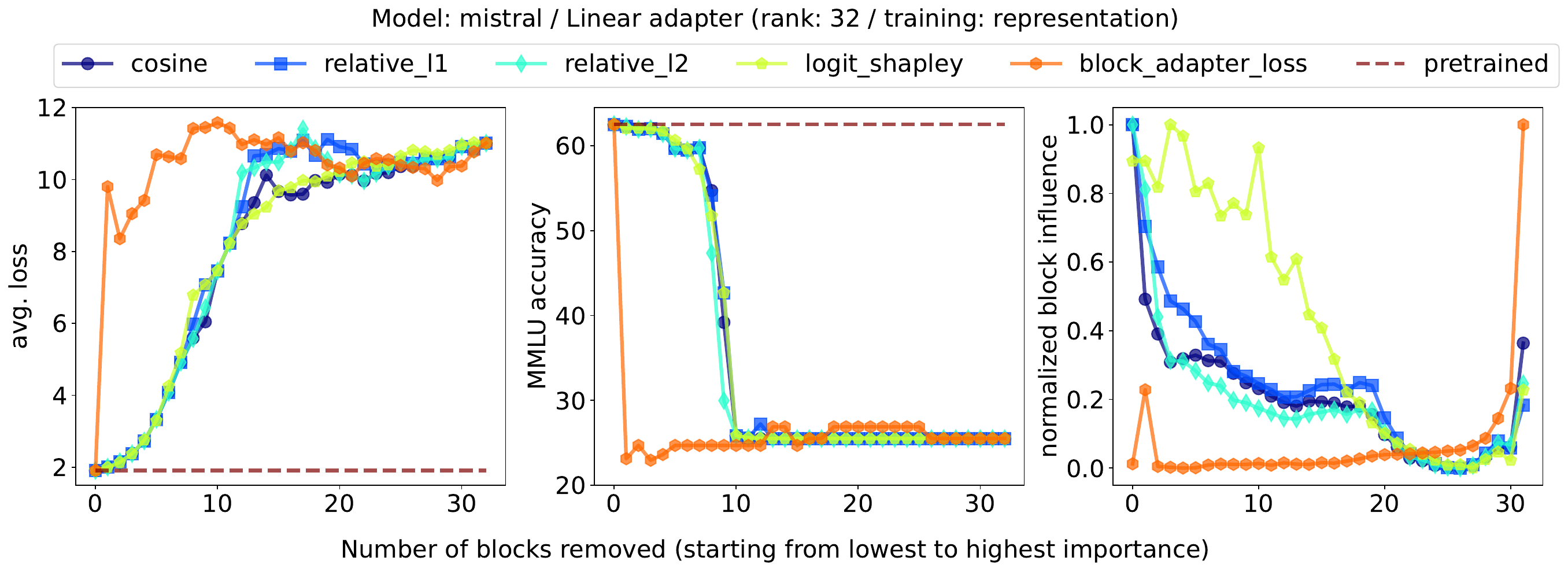}
    }

    \subfloat{
        \includegraphics[width=\textwidth]{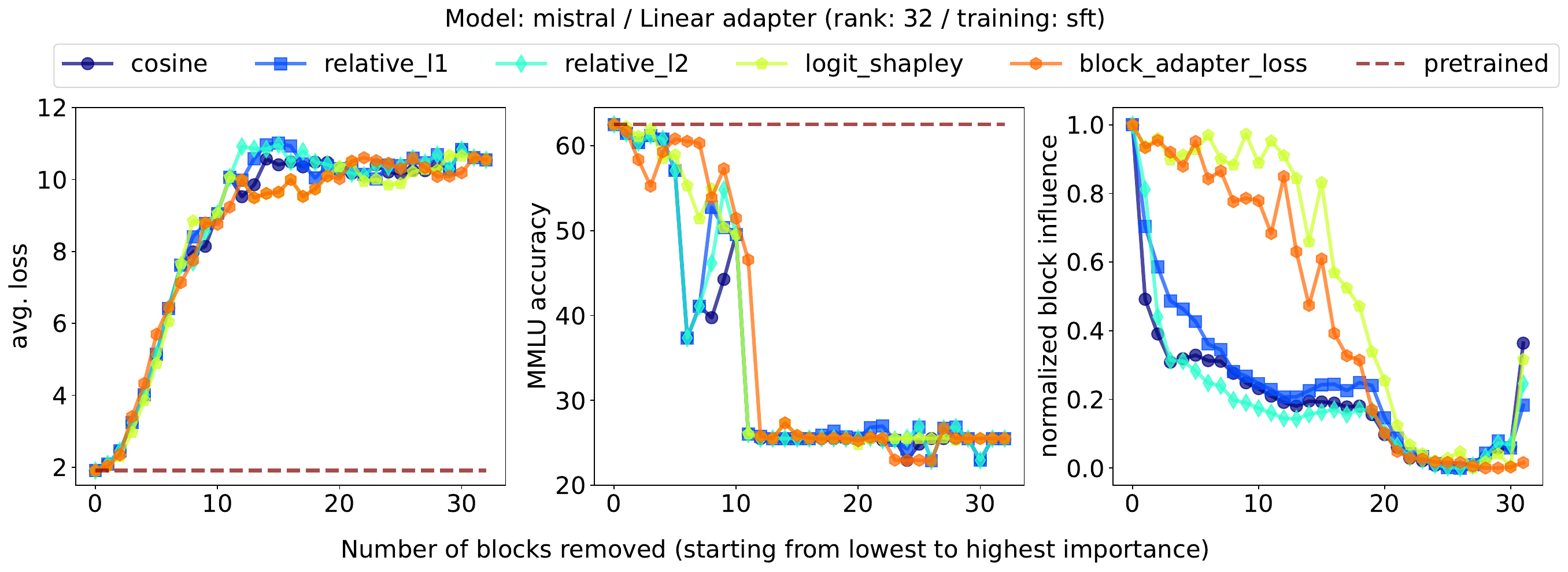}
    }

    \subfloat{
        \includegraphics[width=\textwidth]{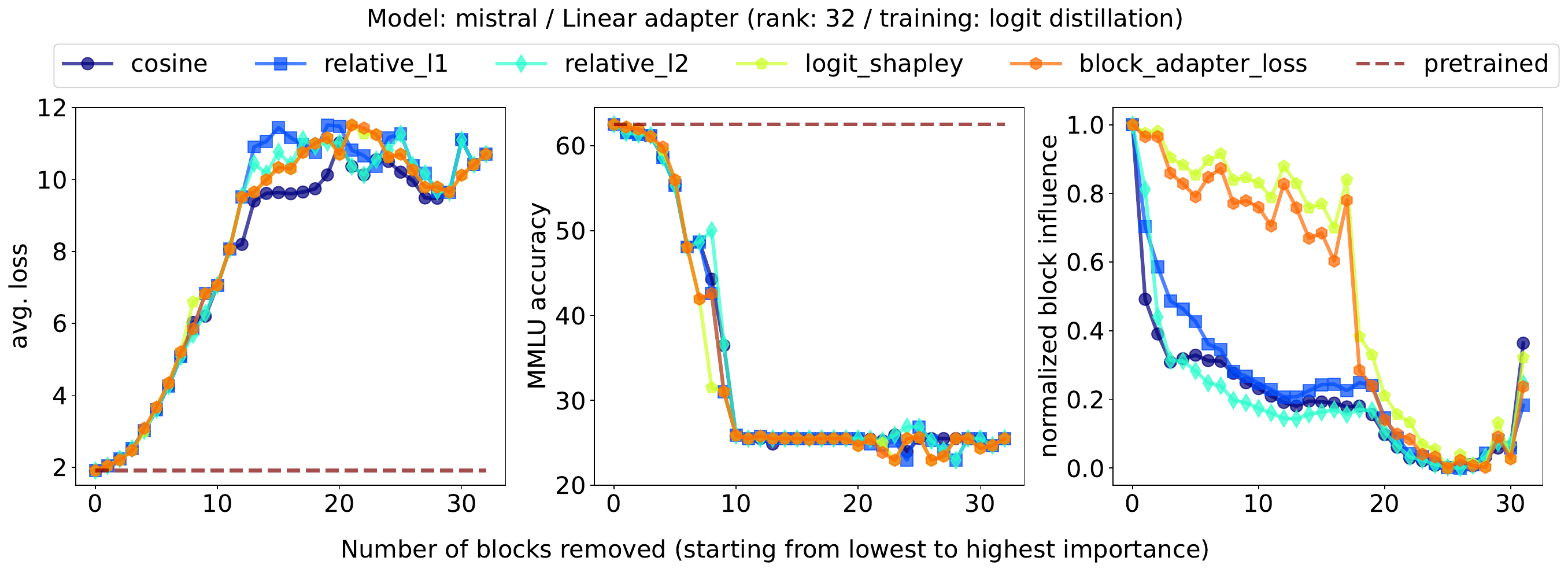}
    }
    \caption{\textbf{Evaluating the impact of linear adapters for Mistral 7b with a rank of 32} trained using three different metrics including (a) MSE loss defined on the representation, (b) supervised fine-tuning (SFT), and (c) logit distillation where logits are distilled from the full model.}
    \label{fig:linear_adapters_rank_32_mistral}
\end{figure}

Results for LLaMa-2 7b and Mistral 7b with an adapter rank of 32 on a relative scale are visualized in Fig.~\ref{fig:linear_adapters_rank_32_llama2_relative} and Fig.~\ref{fig:linear_adapters_rank_32_mistral_relative} respectively.
Results for LLaMa-2 7b and Mistral 7b with an adapter rank of 32 on the original scale are visualized in Fig.~\ref{fig:linear_adapters_rank_32_llama2} and Fig.~\ref{fig:linear_adapters_rank_32_mistral} respectively.

\begin{figure}
    \centering
    \subfloat{
        \includegraphics[width=\textwidth]{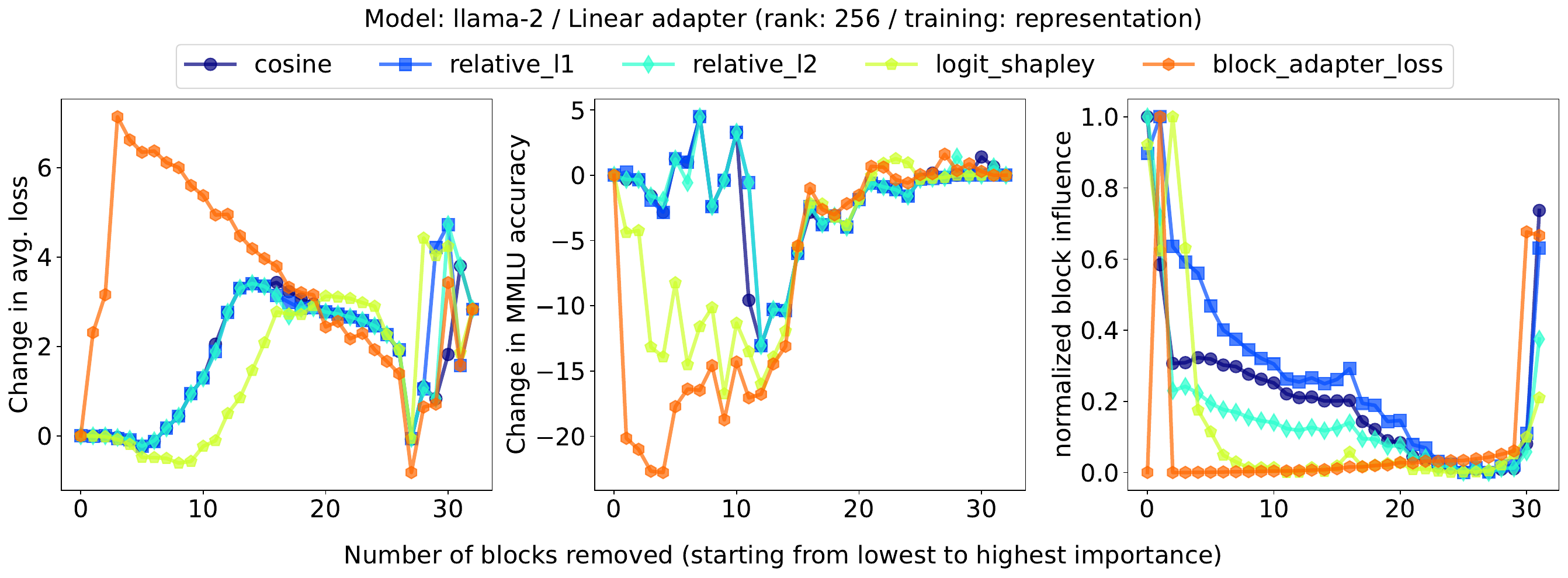}
    }

    \subfloat{
        \includegraphics[width=\textwidth]{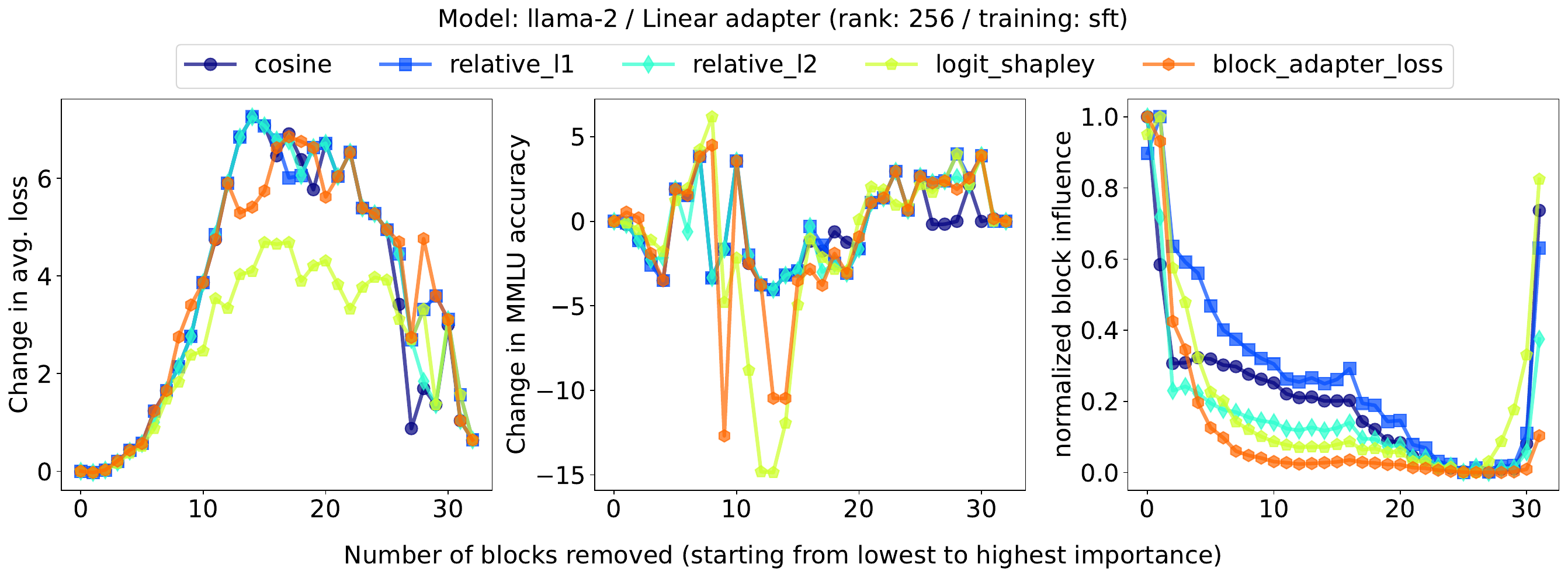}
    }

    \subfloat{
        \includegraphics[width=\textwidth]{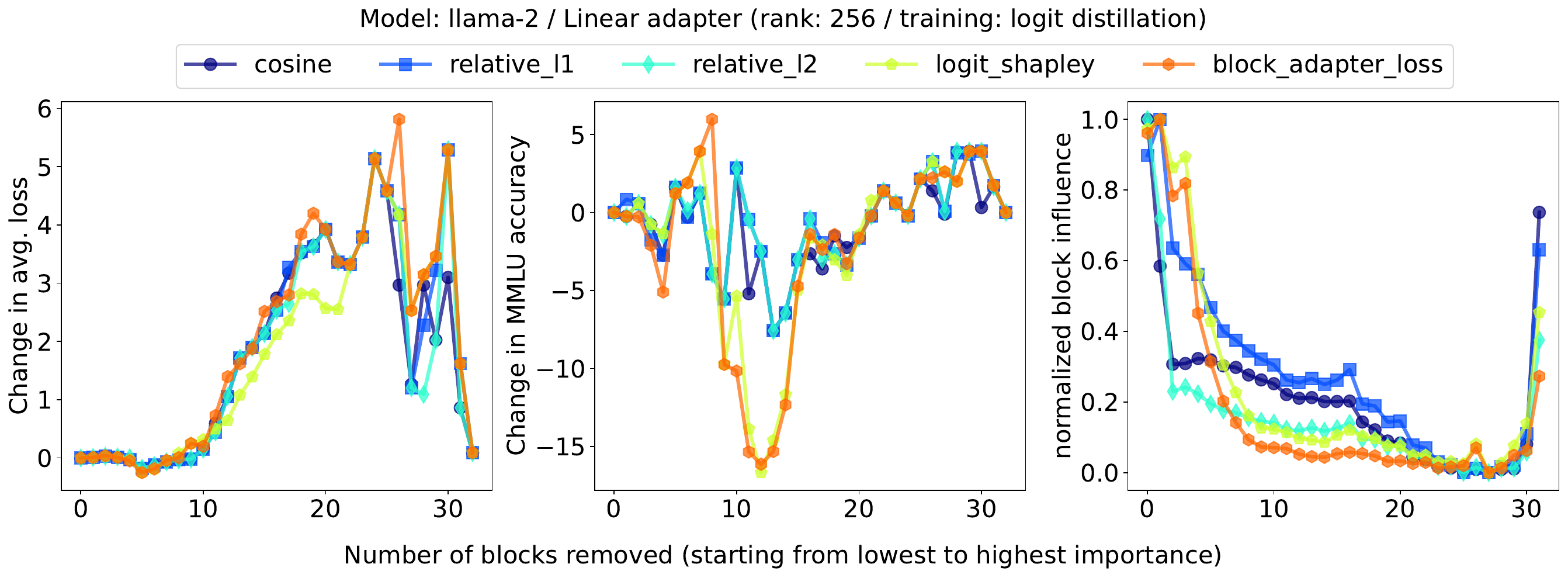}
    }
    \caption{\textbf{Evaluating the relative impact of linear adapters for LLaMa-2 7b with a rank of 256} trained using three different metrics including (a) MSE loss defined on the representation, (b) supervised fine-tuning (SFT), and (c) logit distillation where logits are distilled from the full model. The plot with the original values is presented in Fig.~\ref{fig:linear_adapters_rank_256_llama2}.}
    \label{fig:linear_adapters_rank_256_llama2_relative}
\end{figure}

\begin{figure}
    \centering
    \subfloat{
        \includegraphics[width=\textwidth]{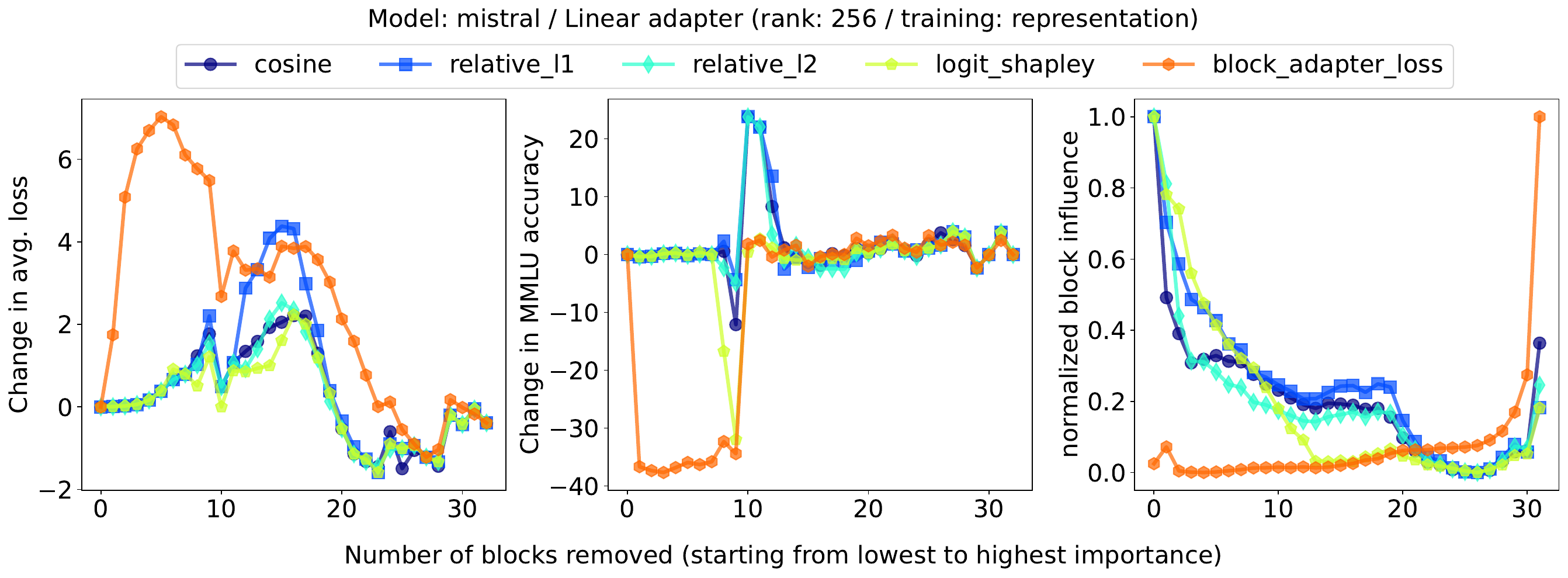}
    }

    \subfloat{
        \includegraphics[width=\textwidth]{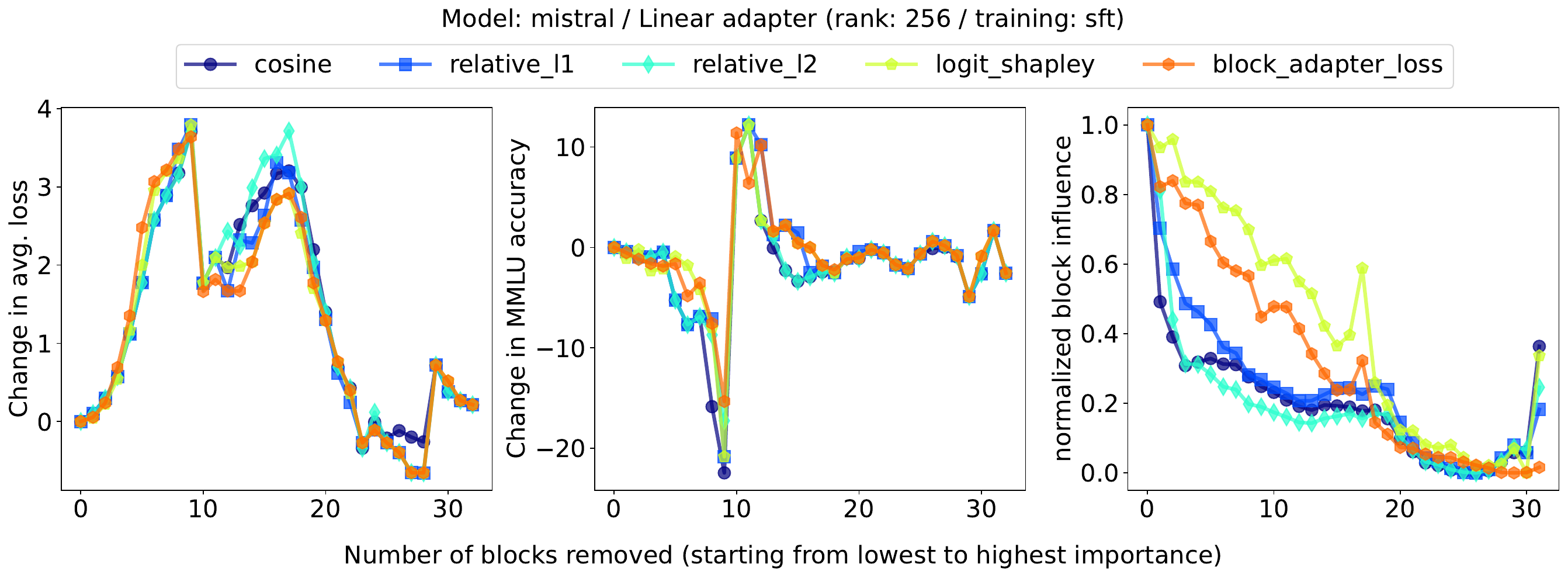}
    }

    \subfloat{
        \includegraphics[width=\textwidth]{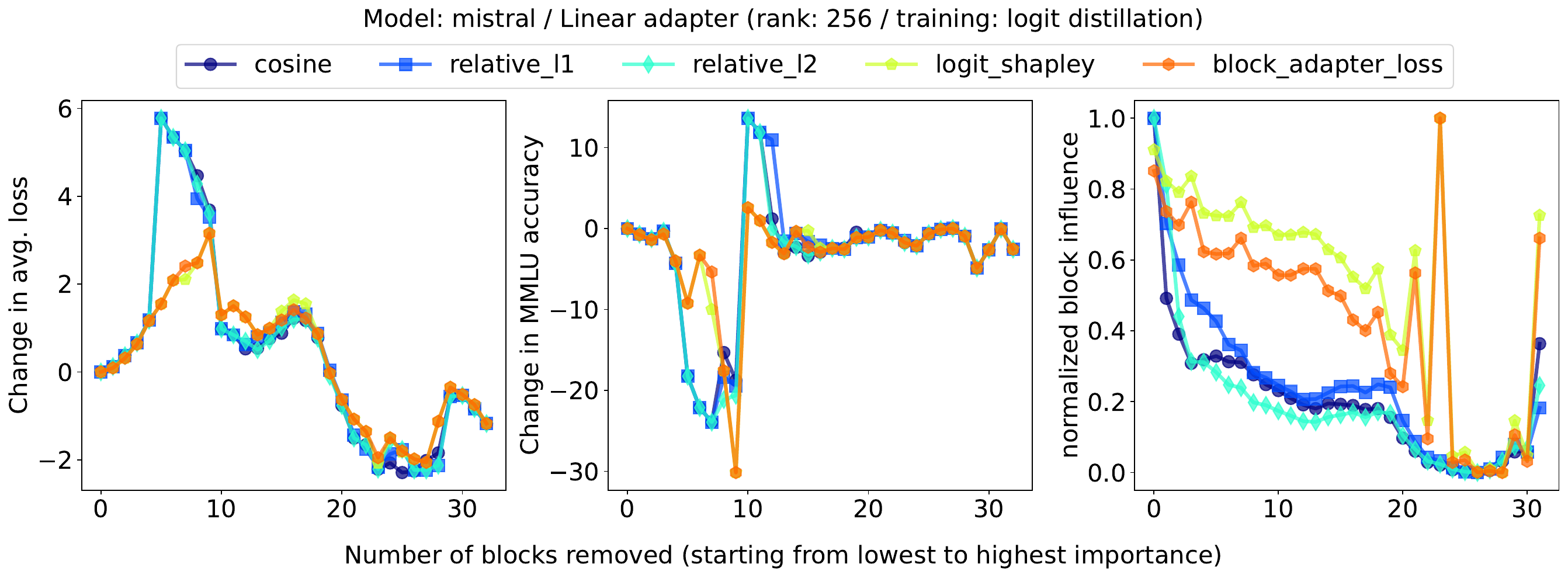}
    }
    \caption{\textbf{Evaluating the relative impact of linear adapters for Mistral 7b with a rank of 256} trained using three different metrics including (a) MSE loss defined on the representation, (b) supervised fine-tuning (SFT), and (c) logit distillation where logits are distilled from the full model. The plot with the original values is presented in Fig.~\ref{fig:linear_adapters_rank_256_mistral}.}
    \label{fig:linear_adapters_rank_256_mistral_relative}
\end{figure}

\begin{figure}
    \centering
    \subfloat{
        \includegraphics[width=\textwidth]{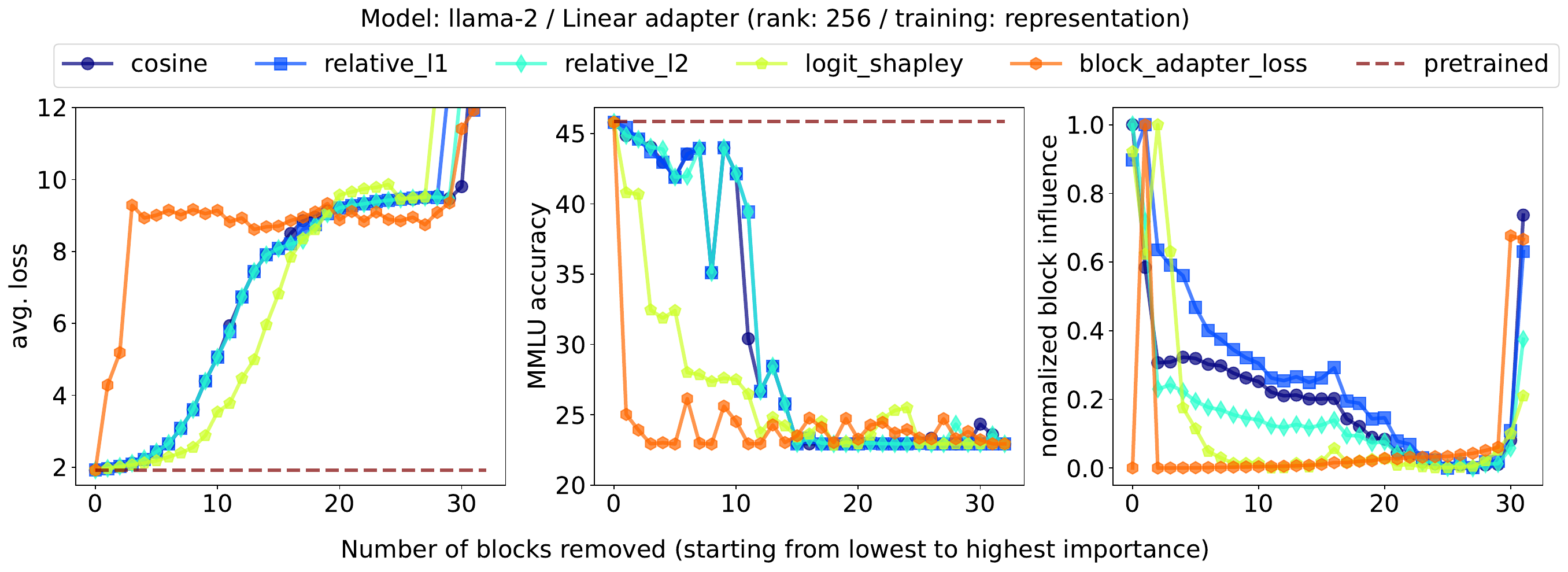}
    }

    \subfloat{
        \includegraphics[width=\textwidth]{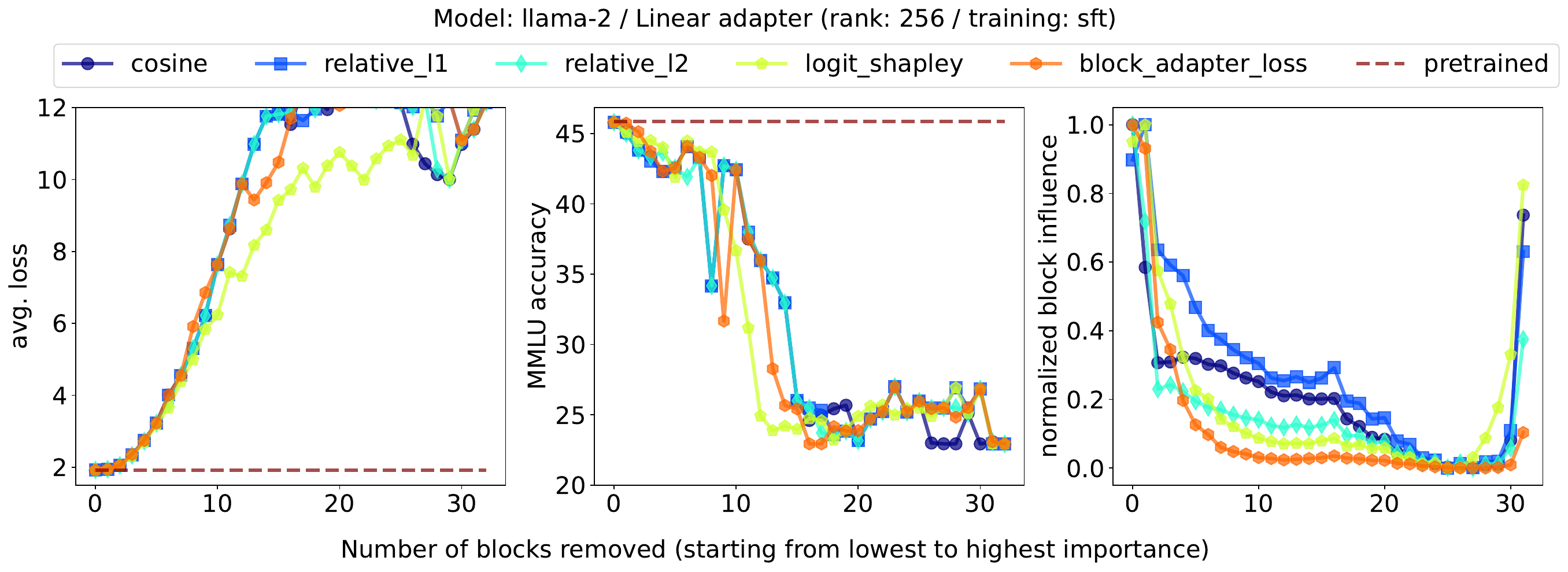}
    }

    \subfloat{
        \includegraphics[width=\textwidth]{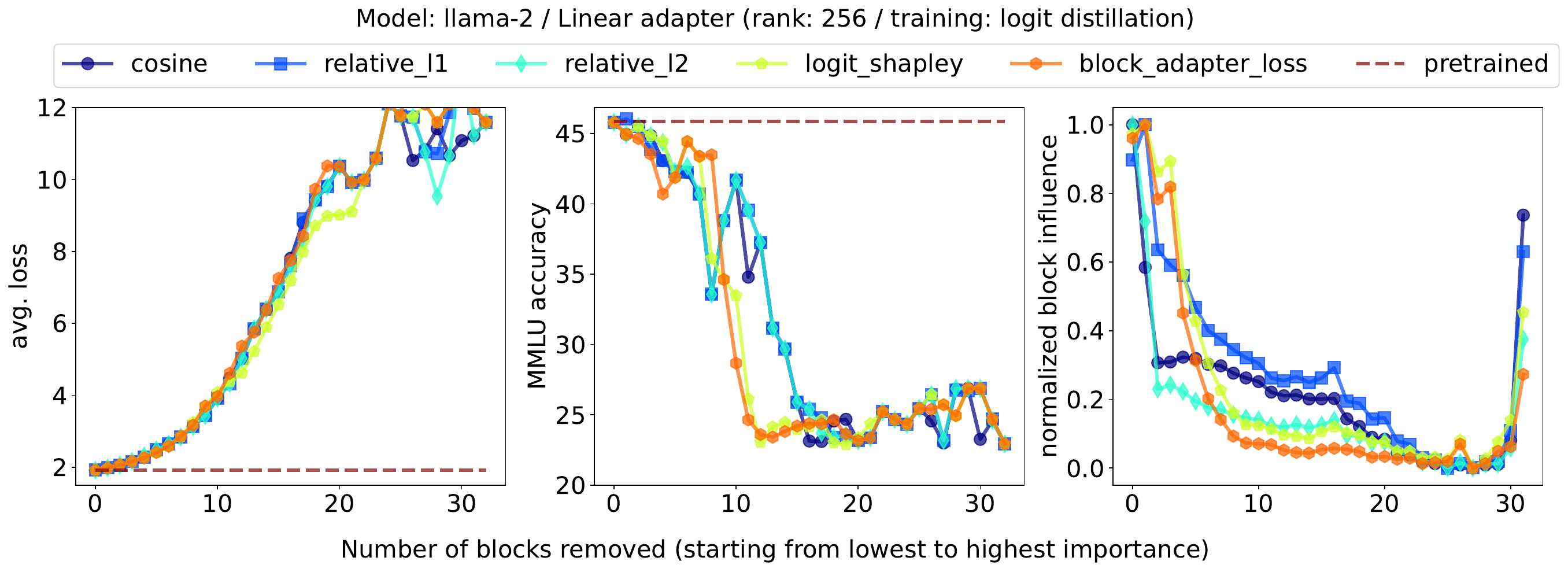}
    }
    \caption{\textbf{Evaluating the impact of linear adapters for LLaMa-2 7b with a rank of 256} trained using three different metrics including (a) MSE loss defined on the representation, (b) supervised fine-tuning (SFT), and (c) logit distillation where logits are distilled from the full model.}
    \label{fig:linear_adapters_rank_256_llama2}
\end{figure}

\begin{figure}
    \centering
    \subfloat{
        \includegraphics[width=\textwidth]{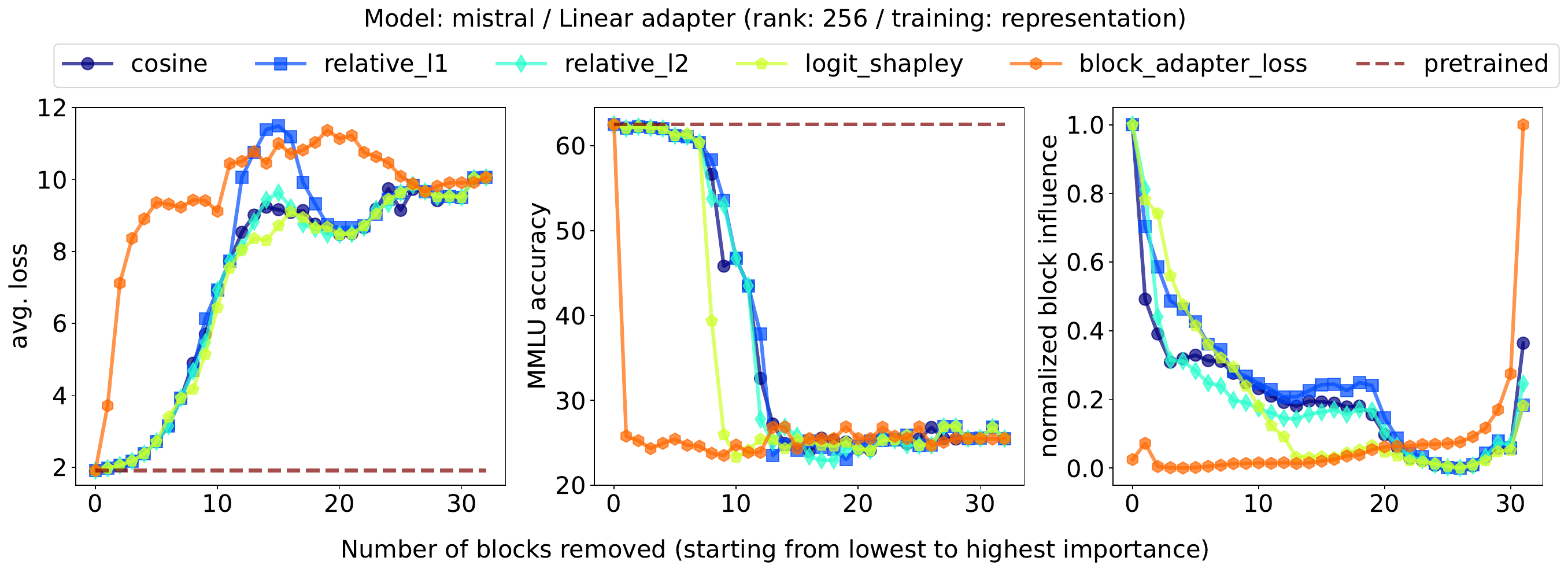}
    }

    \subfloat{
        \includegraphics[width=\textwidth]{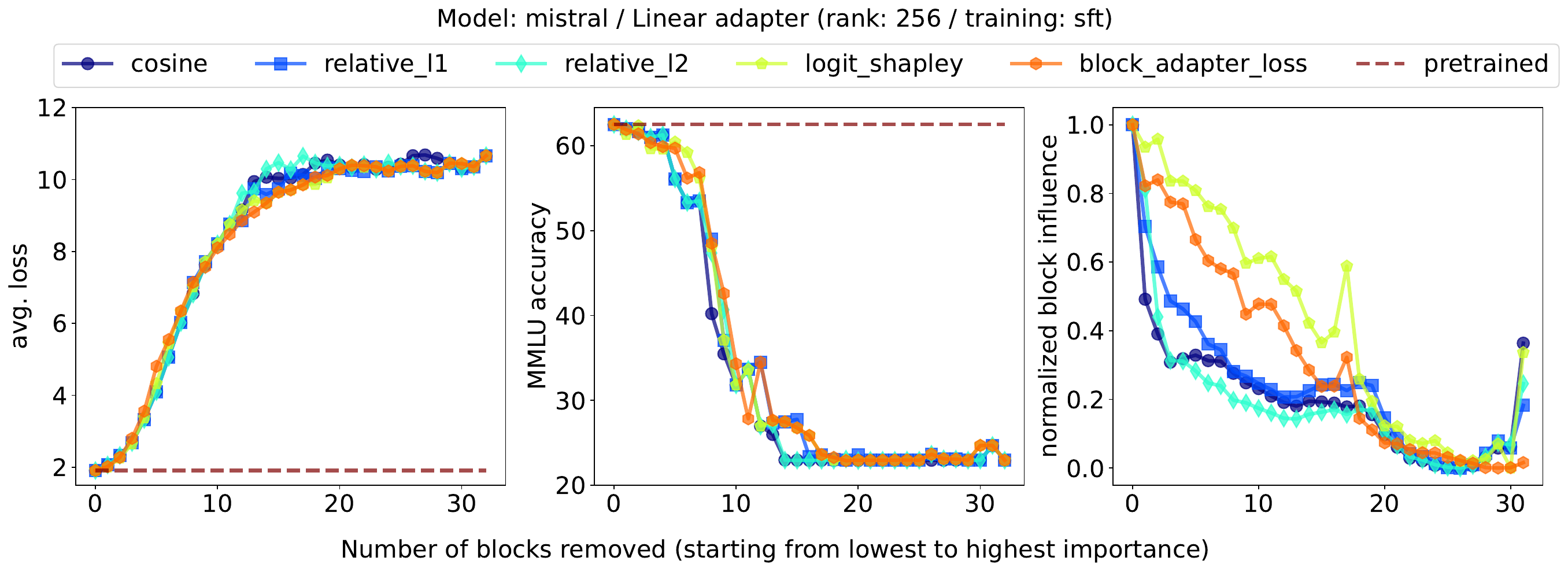}
    }

    \subfloat{
        \includegraphics[width=\textwidth]{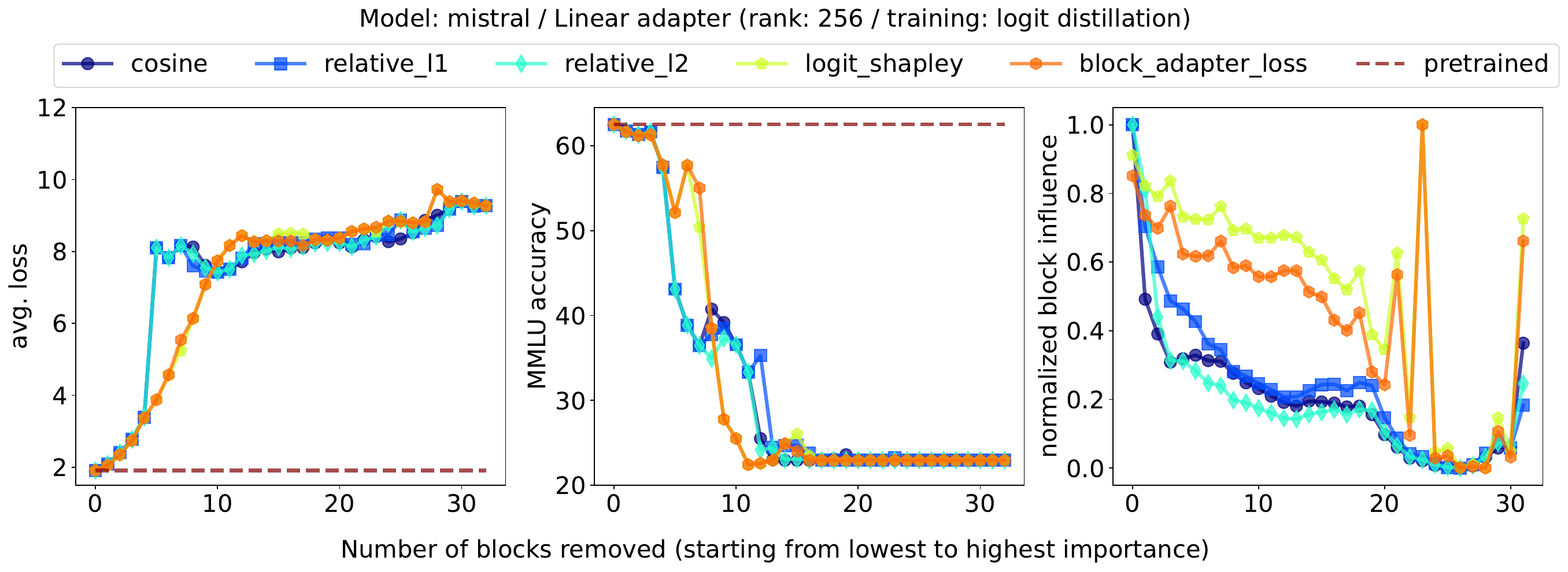}
    }
    \caption{\textbf{Evaluating the impact of linear adapters for Mistral 7b with a rank of 256} trained using three different metrics including (a) MSE loss defined on the representation, (b) supervised fine-tuning (SFT), and (c) logit distillation where logits are distilled from the full model.}
    \label{fig:linear_adapters_rank_256_mistral}
\end{figure}

Results for LLaMa-2 7b and Mistral 7b with an adapter rank of 256 on a relative scale are visualized in
Fig.~\ref{fig:linear_adapters_rank_256_llama2_relative} and Fig.~\ref{fig:linear_adapters_rank_256_mistral_relative} respectively.
Results for LLaMa-2 7b and Mistral 7b with an adapter rank of 256 on the original scale are visualized in
Fig.~\ref{fig:linear_adapters_rank_256_llama2} and Fig.~\ref{fig:linear_adapters_rank_256_mistral} respectively.

\end{document}